\pdfoutput=1

\documentclass[11pt]{article}

\usepackage{acl}

\usepackage{times}
\usepackage{latexsym}
\usepackage{booktabs}
\usepackage{amsmath}
\usepackage{algorithm}
\usepackage{algpseudocode}
\usepackage{graphicx}
\usepackage{subcaption}
\usepackage{multirow}
\usepackage{tabularx}
\usepackage{enumitem}
\usepackage{svg}
\usepackage{tabularray}
\usepackage{longtable}
\usepackage{amsthm}
\usepackage{bm}

\setitemize{labelindent=0em,labelsep=0.2cm,leftmargin=*}

\DeclareMathOperator*{\softmax}{softmax}
\DeclareMathOperator*{\test}{test}
\DeclareMathOperator*{\defence}{def}
\usepackage[T1]{fontenc}

\usepackage[utf8]{inputenc}

\usepackage{microtype}

\usepackage{inconsolata}

\title{Misinformation with Legal Consequences (\texttt{MisLC}): A New Task Towards Harnessing Societal Harm of Misinformation}

\author{Chu Fei Luo\textsuperscript{\hspace{1.2mm}1,2}, Radin Shayanfar\textsuperscript{\hspace{1.2mm}1,2}, Rohan Bhambhoria\hspace{1.2mm}\textsuperscript{1,2}, \\\textbf{Samuel Dahan\textsuperscript{2,3}, and Xiaodan Zhu\textsuperscript{1,2}}\\
\textsuperscript{1}Department of Electrical and Computer Engineering \& Ingenuity Labs Research Institute \\ Queen's University\\
\textsuperscript{2}Conflict Analytics Lab, Queen's University\\
\textsuperscript{3}Cornell Law School\\
\small{\{\texttt{chufei.luo,radin.shayanfar,r.bhambhoria,samuel.dahan,xiaodan.zhu}\}\text{\texttt{@queensu.ca}}}} 

\begin{document}
\maketitle
\begin{abstract}

\textit{Misinformation}, defined as false or inaccurate information, can result in significant societal harm when it is spread with malicious or even innocuous intent. The rapid online information exchange necessitates advanced detection mechanisms to mitigate misinformation-induced harm. Existing research, however, has predominantly focused on assessing veracity, overlooking the legal implications and social consequences of misinformation. In this work, we take a novel angle to consolidate the definition of misinformation detection using legal issues as a measurement of societal ramifications, aiming to bring interdisciplinary efforts to tackle misinformation and its consequence. 
We introduce a new task: Misinformation with Legal Consequence (\texttt{MisLC}), which leverages definitions from a wide range of legal domains covering 4 broader legal topics and 11 fine-grained legal issues, including hate speech, election laws, and privacy regulations. For this task, we advocate a two-step dataset curation approach that utilizes crowd-sourced checkworthiness and expert evaluations of misinformation. We provide insights about the \texttt{MisLC} task through empirical evidence, from the problem definition to experiments and expert involvement. While the latest large language models and retrieval-augmented generation are effective baselines for the task, we find they are still far from replicating expert performance.\footnote{Our code and data are available at \url{https://github.com/chufeiluo/mislc} for replicability.}


\end{abstract}

\begin{figure}
    \centering
    \includegraphics[width=0.95\linewidth]{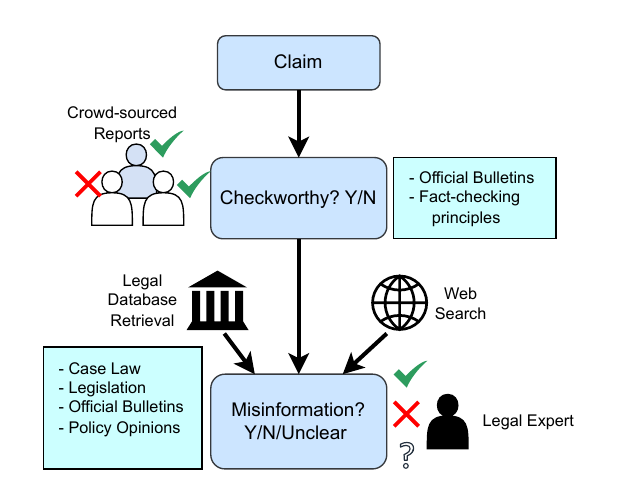}
    \caption{An overview of our proposed task framework for legal misinformation. We obtain crowd-sourced labels of checkworthiness. If a claim is checkworthy, we use legal annotators to annotate potential legal issues of misinformation.}
    \label{fig:pipeline}
\end{figure}
\section{Introduction}
Artificial intelligence is advancing with an unprecedented speed, and many emerging problems with profound societal impact need multi-disciplinary research efforts and solutions. Misinformation, broadly defined as \textit{false} or \textit{inaccurate information,} has had a widespread harmful impact. If unaddressed, it will persist and exacerbate systemic problems in our daily life as well as many critical areas \cite{Budak2024}.
For instance, conflicting information during the COVID-19 pandemic significantly influenced people's attitudes and behaviours toward preventing viral spread \cite{enders2020different}. In significant economic or political events, misinformation also proves extremely detrimental (e.g., in the form of \textit{fake news}), where malicious actors are motivated to purposely spread false information to manipulate public opinion while an event unfolds \cite{doi:10.1080/02650487.2019.1586210}. 




The growing menace 
of online misinformation
underscores the urgent need for regulation, as exemplified by the European Commission's recent action plans.\footnote{\url{https://digital-strategy.ec.europa.eu/en/policies/online-disinformation}} 
We believe that NLP enabled solutions will play a critical role in mitigating the adversarial affects of misinformation. These solutions require a human-centric approach,
with the basic design to ensure the \textit{alignment} between humans and AI, centring on the values and interests of humans \cite{bai2022constitutional, pyatkin-etal-2023-clarifydelphi, dahan2023lawyers}. 
Collaborations with experts in the social sciences are essential to achieve this goal.

In this work, we take a novel angle to define misinformation with its outcome that can be regularized by laws or regulations, building on legal issues as a measurement of societal ramifications, and aiming to bring interdisciplinary efforts to tackle misinformation and its consequence. 
Unlike previous work that has focused on factual accuracy or checkworthiness as potential controversy of a topic \cite{das_state_2023}, 
we ground our definition in legal literature and social consequence.
Our main contributions are summarized as follows:
\begin{itemize}[leftmargin=8pt,itemsep=-0em,before=\vspace{-0cm}]
    \item 
We introduce a new task: Misinformation with Legal Consequence (\texttt{MisLC}), which leverages definitions from a wide range of legal domains covering 4 broader legal topics and 11 fine-grained legal issues, including hate speech, election laws, and privacy regulations. We advocate a two-step dataset curation approach, utilizing crowd-sourced checkworthiness and expert evaluations of misinformation. We expect our process and discussions could help other similar tasks that need to involve costly domain experts to jointly solve problems with significant societal impact.

    \item We evaluate the state of the art of the most recent large language models (LLMs) on \texttt{MisLC}, by performing a comprehensive study on a wide range of  open-source and proprietary LLMs that covers a broad parameter spectrum and varying training data. Two advanced Retrieval-Augmented Generation (RAG) architectures are investigated to detect legally consequential misinformation, involving retrieval from legal document databases and web search, mimicing expert techniques. 
\item  We provide insights about the \texttt{MisLC} task through empirical evidence, from the problem definition to experiments and domain expert involvement. After thorough empirical study, we find the existing LLMs perform reasonably well at the task, achieving non-random performance without external resources. Their performance also increases consistently with RAG. However, LLMs are still far from matching human expert performance. Through this work, we urge further exploration in this challenging task with significant societal impact.

\end{itemize}

\section{Related Work}

Misinformation is a serious issue with significant societal impact, as factual dissonance can cause disorder in peoples' worldviews \cite{doi:10.1080/02650487.2019.1586210}. 
There have been various works that address separate components of the fact-checking pipeline: identifying checkworthy claims, gathering sources on those claims, and predicting veracity \cite{das_state_2023}.
There is growing interest in addressing the problem with LLMs \cite{chen2023combating,bhambhoria-etal-2023-simple}, and emerging works proposing new methodologies for fact-checking \cite{pelrine2023towards, pan-etal-2023-fact}. However, these works do not consider the legal concept of misinformation. 

 
Generative, or auto-regressive models 
have recently demonstrated strong proficiency in a wide variety of tasks such as relevance, stance, topics, and frame detection in tweets \cite{gilardi2023chatgpt, bang2023multitask}. 
Large Language Models (LLMs) have also demonstrated the ability to capture and memorize a vast amount of world knowledge during pretraining \cite{guu2020realm}. However, this knowledge is stored implicitly within their parameters, leading to a lack of transparency for the facts and information generated in their outputs \cite{rashkin2023measuring, manakul2023selfcheckgpt}. One viable strategy for factual accuracy is giving explicit knowledge from external corpora, or Retrieval-Augmented Generation \cite{du-ji-2022-retrieval}. 
Some approaches prepend retrieved documents in the input \cite{guu2020realm,shi2023replug,luo-etal-2023-legally}. 
For further related work, please refer to Appendix \ref{app:drw}.



\section{Misinformation with Legal Consequences (\texttt{MisLC})}

\subsection{Definition}
\label{sec:definition}

The \texttt{MisLC} dataset $\bm{\mathbf{D}} \equiv \{\mathbf{d}_{1}, \dots, \mathbf{d}_{N}\}$ is composed of $N$ instances, each $\mathbf{d}_{i} \in \bm{\mathbf{D}}$ being a tuple $\langle \mathbf{t}_i, \mathbf{E}_i, L_i, y_i \rangle$, where $\mathbf{t}_i$ is a piece of text (e.g., a social media article) represented as a vector of tokens. $\mathbf{E}_i$ is a set of external \textit{evidence} documents that can be used to support or refute the text $\mathbf{t}_i$. $L_i \subset L$ is a subset of \textit{legal issues} from a predefined issue set $L$. Each legal issue refers to an area of law that can be used to indict or punish misinformation, e.g., \textit{Election Laws}, \textit{Public Mischief}, or \textit{Cyberbullying}. We will discuss the details in this section.  

The coarse-grained label of \texttt{MisLC} is $y_i \in \{0, 1, 2\}$, where `$2$' represents \textit{Misinformation with Legal Consequence} (\texttt{MisLC}), `$1$' denotes \textit{Unclear}, and `$0$' denotes the negative class, not \texttt{MisLC} (\texttt{Non-MisLC}).
\textit{Unclear} is reserved for cases that are impossible to determine a classification when there is insufficient context to make the decision. This label is crucial because in real-life applications, we need to separate them for further legal processing, including collecting more evidence. The details will be further discussed in Section \ref{sec:dataset}. The \texttt{MisLC} evaluation is organized in two settings: (1) a binary task, with \texttt{MisLC} as the only positive class of interest, and the other two as negative, and (2) a 3-way classification task, where \texttt{MisLC} and \texttt{Unclear} are separate positive classes.





The evidence $\mathbf{E}_i$ and legal issues $L_i$ are used by legal professionals to obtain the ground-truth labels for \texttt{MisLC}. A necessary condition of a span of text $\mathbf{t}_i$ being \textit{misinformation with legal consequence} is that it makes a claim, where a claim is defined as ``stating or asserting that something is the case, typically without providing evidence or proof.''\footnote{\url{https://languages.oup.com/google-dictionary-en/}} For one legal issue $l \in L$, there is an associated tuple of tests and defenses $(\test_j(\mathbf{t}, \mathbf{E}),\ \defence_j(\mathbf{t}, \mathbf{E}))$. A claim can \textbf{pass} a test or a defense, which we denote with logical True, and failure is denoted by logical False. If a expert annotator assesses a claim in $\mathbf{t}_i$ is associated with a legal issue, i.e. it passes its relevant legal tests and does not pass possible defences, this will trigger the \texttt{MisLC} label. Formally, the set of legal issues $L_i = \{\test_j(\mathbf{t}_i, \mathbf{E}_i)\land\ \neg\ \defence_j(\mathbf{t}_i, \mathbf{E}_i)\ \forall\ l_j \in L\}$, where $l_j$ is a tuple of tests and defences $(\test_j(\mathbf{t}, \mathbf{E}),\ \defence_j(\mathbf{t}, \mathbf{E}))$.  
    \begin{equation}
    \label{eq:labeldef}
        y_i = 
        \begin{cases} 
        2 &  |\mathbf{E}_i| > 0\ \land\ |L_i| > 0\\
        0, &  |\mathbf{E}_i| > 0\ \land\ |L_i| = 0 \lor \mathbf{t}_i~\text{ not a claim}\\
        \end{cases}
    \end{equation}


\paragraph{Legal Resources.} Defining misinformation from a legal standpoint is challenging. Misinformation is an umbrella term to capture the act of publishing any form of false or misleading information in a public space. This is reflected in current legal practices; the issue of false or misleading information may fall under multiple distinct area of law. For example, misinformation aimed at a target group can be punished under hate speech laws. We note that there are very few jurisdictions with provisions that directly address misinformation as a separate legal issue. 
Since the definitions of misinformation are broad, they better serve as an indication of a policy domain rather than a legal category \cite{van2019legal}. Despite concerns about regulating misinformation \cite{o2021perils}, the existing laws have been crafted through extensive discussion to mitigate the harm caused by misinformation to society, reflecting a deliberate and thoughtful approach to a complex issue. 

We collaborate with legal annotators to build a text database on the legal definition of misinformation. Our search spans diverse legal areas, including hate speech, consumer protection, election laws, defamation, food and drug safety, and privacy regulations. For specific citations, please refer to Appendix \ref{app:legal}.
We consider the following sources:
\begin{itemize}[leftmargin=8pt,itemsep=-0em,before=\vspace{-0.1cm}]
    \item \textbf{Legislation} --- Written laws that provide rules of conduct regarding misleading or false information. This encompasses both criminal laws, which can lead to incarceration, and civil laws, which can result in fines.
    \item \textbf{Case Law} --- The written decisions of judges in higher court cases that have been made on misinformation issues. We focus on Canadian and European cases.
    \item \textbf{Official Bulletins} --- Publications from global organizations, including Unicef, the European Union, the United Nations High Commissioner for Refugees (UNHCR), and the Canadian Government, regarding the general definition and identification of misinformation.
    \item \textbf{Policy Opinions} --- Publications from reputable policy makers on misinformation and how legal policy should be applied to prevent its harm. 
\end{itemize}
\paragraph{Legal Issues.} From the legal resources
, we compile 11 major \textbf{legal issues} to form $L$. The full set is listed in Appendix \ref{app:legal}. Each issue has two components: a \textbf{legal test} to determine potential violations, and \textbf{defences} that can counter allegations 
. Our legal issues $l \in L$ are mostly differentiated by the topic/nature of the post, as well as a contention between the post's intent, consequences, and truthfulness. One defence is proving a statement as fact --- if the actor can establish their statement is true, their post is no longer punitive. However, this largely depends on the legal issue. The \textit{defamation} issue, for example, focuses on the \textit{defamatory} nature that has the ability to lower someone's reputation, as well as the context the claim was uttered. All relevant legal tests $\test(\mathbf{t}, \mathbf{E})$ and defences $\defence(\mathbf{t}, \mathbf{E})$ were compiled into a comprehensive annotation guide.

\subsection{Creation of the \texttt{MisLC} Dataset}
\label{sec:dataset}
As illustrated in Figure \ref{fig:pipeline}, we advocate a two-stage data curation process. First, non-legal crowd-sourced annotators discover checkworthy misinformation samples that arouse their suspicion. Second, legal experts annotate the samples and decide relevant legal issues. 
\paragraph{Crowd-sourced Checkworthiness.}
We first want to utilize a layperson's ability to identify misinformation. This component does not require legal expertise, but builds the dataset on which legal practitioners can operate. 
We sampled social media data from \citep{chen2023tweets}, a large public domain dataset with Twitter data (tweets) regarding the \emph{Russia-Ukraine conflict}. We choose the Russia-Ukraine conflict as a recent event with a significant amount of misinformation, and is extensively studied in previous works \citep{alyukov-etal-2023-wartime,tracey-etal-2022-study}. For more details on data processing and data samples, please refer to Appendix \ref{app:data}.

We collected crowd-sourced annotations on this data for \emph{checkworthiness,} in order to filter samples that are likely to contain legal consequences. The crowd-sourced workers could choose between three labels: Checkworthy, Not Checkworthy, or Invalid/No Claim. We sourced our definition of checkworthiness from \citep{das_state_2023}. Additionally, we incorporated indicators of disinformation from an official bulletin released by the Government of Canada. \footnote{\url{https://www.canada.ca/en/campaign/online-disinformation.html}} To ensure data quality, we conducted a pre-screening test with a pool of 100 samples using the same instructions as the main task. This pool was labelled by two members of the research team given the annotation instructions. Details of annotator instructions and the prescreening procedure are contained within Appendix \ref{app:annotation}. After this screening process, we obtained a pool of eleven Turk workers for annotations. We randomly sample an additional 1,500 tweets from the 4,000 that we had collected, and provided these to our Turk workers in batches of 500 over one month. 
\paragraph{Adversarial Filtering.}
We performed a secondary adversarial data filtering step to ensure the data is sufficiently consistent. Compared to previous works \citep{sakaguchi2021winogrande}, we replaced cross-entropy loss with KL divergence over the annotation distribution to model annotator disagreement. We score each sample by its training loss as defined in Equation \ref{eq:loss} and Algorithm \ref{alg:filtering}. We perform this filtering three times with $k=1000$ and retain a set of 711 samples that is consistently kept in each trial. The filtering process biases the label distribution to Checkworthy samples, as shown in Table \ref{tab:sources}. This complements our intended pipeline where a sample is flagged by laypeople and further investigated by legal annotators, and indicates strongly Checkworthy samples are likely more consistent than ambiguous agreement. Further details are discussed in Appendix \ref{app:filtering}.
\begin{equation}
\label{eq:loss}
    \mathcal{L}(\mathbf{t}_i, y_i) = D_{KL}(f(\mathbf{t}_i), \softmax(y_i))
\end{equation}

\begin{algorithm}[t]
\caption{Our adversarial filtering process.}
\label{alg:filtering}
\renewcommand\arraystretch{0.75}
\begin{algorithmic}[1]
\Require Dataset $\{\mathbf{t}_i, y_i\}^n_{i = 1} \in X$, target dataset size $k$
\State Encode all samples as the last embedding layer $f(\mathbf{t}_i) = E_{LM}(\mathbf{t}_i)[-1]$
\State Apply softmax to all $y_i \in X$
\State Initialize $X' = X$
\While{True}
    \State Train a linear classifier $f(t)$ on $X'$
    \For{$(\mathbf{t}_i, y_i) \in X'$}
        \State $s_i = \mathcal{L}(\mathbf{t}_i, y_i)$
    \EndFor
    \State $\tau_\mu = \mu(score)$
    \State $S = \{(\mathbf{t}_i, y_i) \in X' | s_i > \tau_\mu\}$
    \If{$|X'\ \backslash\ S| < k$}
        \State \textbf{break}
    \Else
        \State $X' = X'\ \backslash\ S$
    \EndIf
\EndWhile
\end{algorithmic}
\end{algorithm}


\paragraph{Legal Annotations.}
We collaborated with eleven law researchers to annotate our dataset. The law researchers are graduate students in their first and second year
. They performed 2-3 hours of annotations per week as part of a practicum course, and received credits as compensation. Each expert is provided a document summarizing the legal tests and defences $\in L$. The legal experts first determined whether any claims in a sample, or the sample in its entirety, qualify as misinformation by selecting one of the following three options: \texttt{yes}, \texttt{no}, or \texttt{unclear}.
After this preliminary step, the legal experts identify whether the sample raises any potential legal issues. If it does not, the annotators can then specify whether this is due to an available defence (\texttt{defence}) or a lack of factual claims (\texttt{noClaim}). Each sample is annotated three times, and we obtain an overall label via majority voting. We also decide the legal issues $L_i$ for a sample $\mathbf{t}_i$ via majority voting. The nominal Krippendorff's Alpha for the legal annotators is 0.441, while the minimum recommended threshold for reliable data is 0.667 \cite{krippendorff2018content}. However, this Krippendorff's Alpha is consistent with previous works in legal task datasets \cite{thalken2023modeling}. This indicates greater subjectivity in legal tasks, possibly due to their complexity and opportunity for interpretation.

\begin{figure}[t]
    \centering
    \includegraphics[trim={0.5cm 0.75cm 0.5cm, 5.5cm},clip, width=0.9\linewidth]{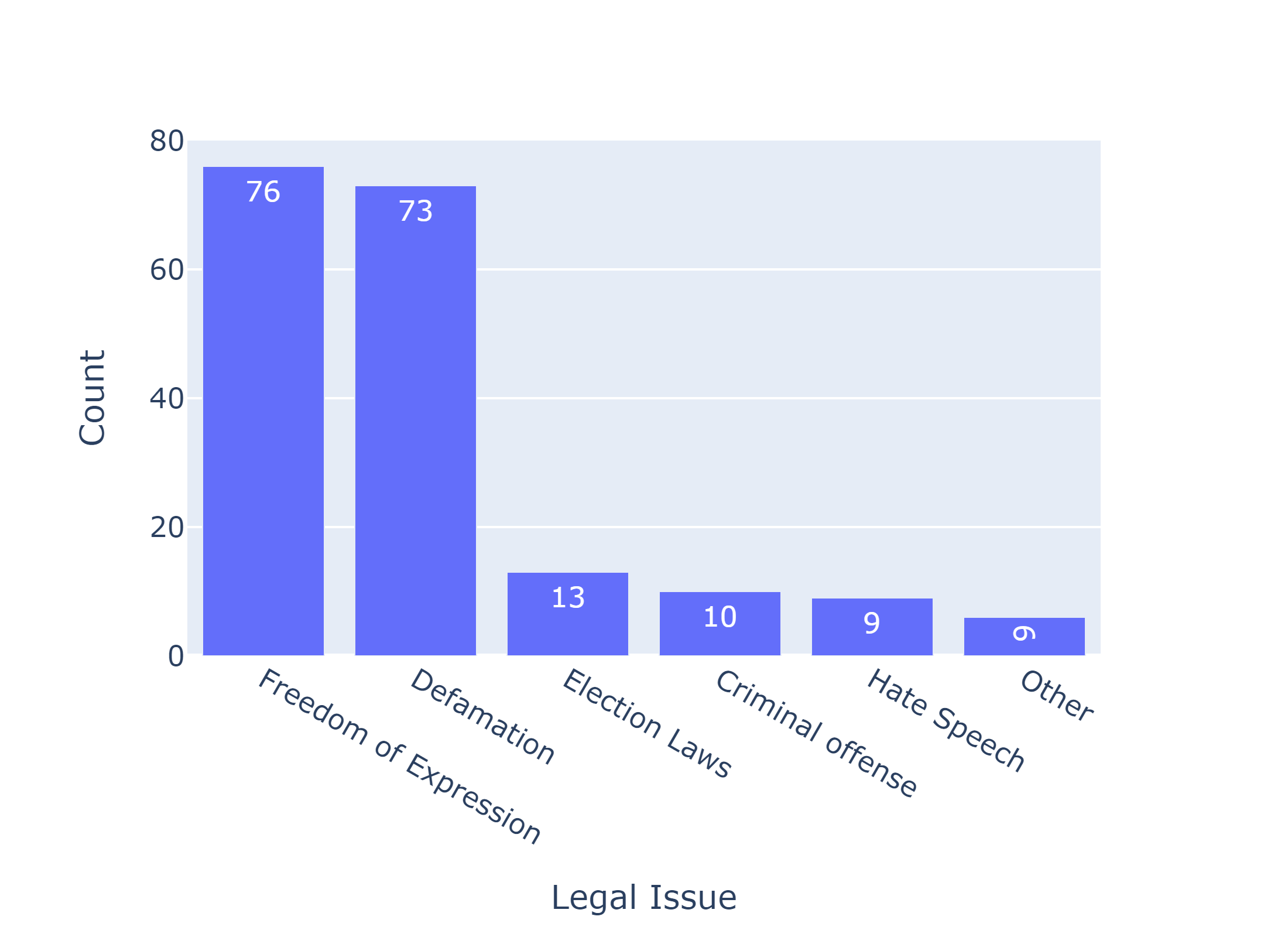}
    \caption{The most frequent legal issues that appear in our dataset.}
    \label{fig:legalissues}
\end{figure}

\begin{table}[t]

\renewcommand\arraystretch{0.5}

    \centering

    \setlength{\tabcolsep}{3pt}

    \resizebox{0.6\linewidth}{!}{

    \begin{tabular}{p{4cm}|c}

    \toprule

        Label & Count \small{(\%)} \\ 
        \midrule

        Total Dataset Size & 711 \small{(100)}   \\
        \midrule
        \texttt{Checkworthy}     &  650 \small{(91.4)} \\
        \midrule
        \texttt{MisLC} &   93 \small{(13.1)} \\
        \texttt{Non-MisLC} & 540 \small{(75.9)} \\
        \texttt{Unclear} & 78 \small{(11.0)} \\
        \midrule
        \texttt{Evidence available}  & 363 \small{(51.0)}  \\
        \midrule
        \texttt{noClaim}  & 304  \small{(42.8)}  \\
        \texttt{defence} & 242 \small{(34.0)} \\
         \bottomrule

    \end{tabular}

    }

        \caption{Statistics on our dataset, including total dataset size, the number of crowd-sourced checkworthy samples, label distribution for \texttt{MisLC}, and special labels from legal annotations. Each sample can have one ground truth label (\texttt{MisLC}, \texttt{Non-MisLC}, or \texttt{Unclear}). }  

    \label{tab:sources}

\end{table}
As shown in Figure \ref{fig:legalissues}, the most relevant legal issues for our data to be Freedom of Expression, followed by closely by Defamation. Next, there are Election laws, the criminal offenses of Cyberbullying and Public Mischief, and Hate Speech. Our label distribution is summarized in Table \ref{tab:sources}. While checkworthiness had a positive rate of 91.4\% (650), only 13.1\% (93 samples) of the dataset has some possible legal violation for misinformation. Additionally, there were a substantial number of \texttt{Unclear} samples (11.0\%, or 78). These are samples with unclear context or implications that annotators felt could not be fact-checked, e.g. ``we all know what he did.'' In the context of our formal definition, this implies the evidence $E_i$ is non-existent, or $|E_i| = 0$. Examining the samples that were checkworthy but not a legal violation, there are a few recurring themes:

\textit{The claim is supported by a reputable source after fact-checking.} We explicitly instructed the crowd-sourced workers to ignore truthfulness of a statement, so this is an expected outcome. This also demonstrates the importance of identifying $E_i$.

    \textit{The claim was deemed to be an opinion.} A key component of the crowd-sourced annotator instructions, sourced from a bulletin by the Canadian Government on disinformation, was whether or not a claim invoked an emotional reaction. There appears to be a subtle distinction between an outrageous claim and a personal/political opinion not captured in the crowd-sourced annotations. After manually inspecting some annotations, we found that the annotators sometimes did not investigate a claim if it was combined with opinion.
\paragraph{Human Expert Performance.} We also calculate an approximation of human performance on our task as an upper bound. First, we assign a random number to each annotator and retrieve all of their individual annotations. For statistical significance, we only retain experts that performed more than 50 annotations. Next, treating their annotations as predictions, we calculate their performance against the majority vote label. We include the mean expert performance for reference in Table \ref{tab:ref_perf}.

\section{Models}
\subsection{Experimental Setup}

Along with our dataset, we present a comprehensive set of baselines evaluating the performance of state-of-the-art LLMs on detecting misinformation with legal consequences.
We examine a wide range of both proprietary and open-source LLMs:
\texttt{GPT-4o}\footnote{\url{https://openai.com/index/hello-gpt-4o/}}, \texttt{GPT-3.5-turbo}~\cite{ouyang2022training}, \texttt{Llama2-(7b, 13b, 70b)}~\cite{touvron2023llama}, \texttt{Llama3-(8b, 70b)}\footnote{\url{https://ai.meta.com/blog/meta-llama-3/}}, \texttt{Mistral-7b}~\cite{jiang2023mistral}, and
\texttt{Solar-10b}~\cite{kim2023solar}. We choose Llama 2 and 3 to isolate the effect of model size, since these suites of models are trained with the same method at various parameter counts. We compare this suite to three open-source models trained on various combinations of fine-tuning, instruction tuning, and Proximal Policy Optimization (PPO) or Direct Policy Optimization (DPO). The \texttt{Solar-10b} we test combines two checkpoints of Solar \citep{kim2023solar}: Solar-Instruct, trained with instruction tuning, and OrcaDPO. Please refer to Appendix~\ref{app:models} for further details on the models and Appendix~\ref{app:settings} for additional experimental details, including hyperparameters and prompt templates.
\paragraph{Evaluation Method.}
The models are first prompted to classify misinformation based on $\mathbf{t}_i$ without any external knowledge, purely based on their understanding of misinformation along with some evidence $E_i$ potentially available in their parametric knowledge. Intuitively, this should be equivalent to the crowd-sourced annotators, and \textit{we do not expect good performance.} 
Our prompt template can be found in Appendix~\ref{app:prompting}.
We ask the model to only output one of three keywords: ``\texttt{misinformation}'' for \texttt{MisLC}, ``\texttt{factual}'' for \texttt{not MisLC}, or ``\texttt{unsure}'' for \texttt{Unclear}. Then, we search the generated text for one of these keywords. If none of these keywords are present, we count the generation as an \textit{error}, and report Error Rate (ER) for each model. Errors are converted into a \texttt{Not MisLC} prediction, i.e. label 0. We also report Binary F1 (Bin-f1) as performance in the binary task setting, and Macro- and Micro-f1 (Ma-f1, Mi-f1) for 3-way classification.


\subsection{Retrieval Enhanced Pipeline}
LLMs have a significant amount of world knowledge, but our task of misinformation with legal consequences relies on legal material that likely does not exist in their pre-training data. As discussed in Section \ref{sec:definition}, our ground truth labels are not just determined by the input text $\mathbf{t}_i$, but also the relevant legal issues $L_i$ and evidence $\mathbf{E}_i$. We use RAG to introduce knowledge from our legal literature, as well as to retrieve potential evidence via web search, in order for the model to receive the same information as our legal annotators.
\paragraph{RAG Methods.} We employ a retrieval-augmented approach for our misinformation detection pipeline. Generally speaking, given a document corpus $\mathcal{C}$ and a retrieval system $\mathcal{R}_{\mathcal{C}}$ that can retrieve most related documents to the input query $q$ from corpus $\mathcal{C}$, RAG can be formulated as $p(w_1, ..., w_n) = \prod_{i=1}^{n} p\left(w_i|w_{<i},\mathcal{R}_{\mathcal{C}}\left(w_{<i}\right)\right)$, where $w_{<i}$ is the sequence of tokens preceding $w_i$, i.e. $\mathbf{t}_i$ \cite{ram-etal-2023-context}. In this work, we experiment with two state-of-the-art RAG methods.
We choose these methods as they do not require pretraining or fine-tuning LMs, which can be expensive due to the large LM sizes. These methods also do not require access to the LMs layers and weights.

In-Context RALM (\textbf{\textit{IC-RALM}}) \cite{ram-etal-2023-context} uses the given input $w_{<i}$ as a query to retrieve a document, and prepends the document to the prompt to generate the output. In this approach, the retrieval is triggered at fixed generation intervals, or retrieval strides $\delta$. To avoid information dilution with long queries, the query is limited to the last $\ell$ tokens of the $w_{i}$. More formally, IC-RALM is formulated in Equation \ref{eq:ralm}, where $q_j^{\delta,\ell} = w_{\delta.j-\ell+1}, ..., w_{\delta.j}$ and $[a; b]$ denotes the concatenation of strings a and b.

    \begin{equation}
    \label{eq:ralm}
    \begin{split}
        p(w_1, ..., w_n) &= \\
        \prod_{j=0}^{n_\delta-1}\prod_{i=1}^{\delta} &p\left(w_{\delta.j+i}|\left[\mathcal{R}_{\mathcal{C}}(q_j^{\delta,\ell}); w_{<i}\right]\right)
    \end{split}
    \end{equation}
    
\textbf{\textit{FLARE}} \cite{jiang-etal-2023-active} generates a temporary sentence $\hat{s}$, where $p(\hat{s}) = \prod_{i=1}^{m} p\left(w_i|w_{<i}\right)$, and then chooses whether to regenerate the sentence with retrieval based on model confidence, i.e. the minimum token probability in the sentence. This is formulated in Equation \ref{eq:flare}, where $\theta$ is the threshold parameter.
Moreover, FLARE formulates the regenerated sentence $s'$ as
    $p(s') = \prod_{i=1}^{m} p\left(w_{i}|\left[\mathcal{R}_{\mathcal{C}}(\mathrm{qry}(w_{<i})); w_{<i}\right]\right)$. The query formulation function $\mathrm{qry}(\cdot)$ generates retrieval queries based on the lowest confidence token spans and by masking low confidence tokens. We adapt their implementation to share the same BM25 indexing and retrieval as IC-RALM. Please refer to Appendix \ref{app:retrieval} for further implementation details.
    \begin{equation}
    \label{eq:flare}
        s = 
        \begin{cases} 
        \hat{s} & \text{if all tokens of } \hat{s} \text{ have probs } \geq \theta \\
        s' & \text{otherwise}
        \end{cases}
    \end{equation}

\begin{table*}[t]

\renewcommand\arraystretch{1.2}

    
    \centering
    \setlength{\tabcolsep}{1.5pt}

    \resizebox{1\linewidth}{!}{

    \begin{tabular}{p{2.5cm}||cccc|cccc|cccc|cccc|cccc}

    \toprule

        \multirow{2}{*}{Model}  & \multicolumn{4}{c|}{No Retrieval}&\multicolumn{4}{c|}{IC-RALM (Legal)}& \multicolumn{4}{c|}{FLARE (Legal)} &\multicolumn{4}{c|}{FLARE (Web)}& \multicolumn{4}{c}{FLARE (Legal+web)}\\

        \cmidrule{2-21}
        &Bin-f1$\uparrow$ &Ma-f1$\uparrow$  & Mi-f1 $\uparrow$ & ER $\downarrow$&Bin-f1$\uparrow$ & Ma-f1 $\uparrow$ & Mi-f1 $\uparrow$  & ER $\downarrow$&Bin-f1$\uparrow$  &Ma-f1 $\uparrow$ & Mi-f1 $\uparrow$ & ER $\downarrow$&Bin-f1$\uparrow$  &Ma-f1$\uparrow$  & Mi-f1 $\uparrow$ & ER $\downarrow$&Bin-f1$\uparrow$ &Ma-f1$\uparrow$  & Mi-f1 $\uparrow$ & ER $\downarrow$\\

        \midrule

        \texttt{GPT-3.5-trb}     &    30.4 & 19.3 & 45.8 & 0.0 & 24.1 & 12.0 & 39.8 & 0.0 & 29.7 & 16.1 & \textbf{48.7} & 0.0 & 30.5 & 17.6 & \textbf{49.1} & 0.0 & 31.1 & 17.9 & \textbf{49.6} & 0.0  \\ 
        \texttt{GPT-4o}            &    28.7 & 23.2 & 43.5 & 0.0 & \textbf{35.8} & \textbf{28.5} & \textbf{50.9} & 0.0 & 32.3 & \textbf{25.8} & 46.7 & 0.0 & \textbf{34.5} & \textbf{26.2} & 47.4 & 0.0 & \textbf{37.7} & \textbf{28.0} & 46.7 & 0.0 \\ 
        \midrule
        \texttt{Mistral-7b}        &    27.9 & 17.2 & 42.8 & 6.8 & 25.9 & 19.1 & 39.3 & 0.0 & 24.5 & 21.2 & 41.0 & 0.0 & 21.7 & 21.6 & 44.7 & 0.0 & 16.7 & 18.3 & 42.2 & 0.0 \\ 
        \texttt{Llama2-7b}         &    21.0 & 11.7 & 34.7 & 24.1 & 23.1 & 11.6 & 38.8 & 0.0 & 23.2 & 12.8 & 38.8 & 0.0 & 22.9 & 11.5 & 38.6 & 0.0 & 23.2 & 12.8 & 38.8 & 0.0 \\ 
        \texttt{Llama3-8b}         &    30.7 & 18.0 & 48.2 & 0.0 & 0.0 & 9.9 & 38.8 & 0.0 & 27.2 & 13.6 & 43.0 & 0.0 & 31.1 & 18.0 & 48.1 & 0.0 & 25.3 & 13.8 & 41.6 & 0.0 \\ 
        \midrule
        \texttt{Solar-10b}         &    27.7 & 14.9 & 31.1 & 32.5 & 28.6 & 22.8 & 39.1 & 3.8 & 27.1 & 21.7 & 41.8 & 1.7 & 32.6 & 22.9 & 44.5 & 4.4  & 28.5 & 21.3 & 40.8 & 2.7 \\ 
        \texttt{Llama2-13b}        &    22.0 & 11.0 & 31.8 & 56.1 & 21.7 & 17.6 & 39.6 & 0.1 & 22.4 & 17.3 & 38.9 & 0.0 & 23.4 & 19.2 & 39.0 & 0.0 & 23.0 & 15.6 & 39.0 & 0.0 \\ 
        \midrule
       \texttt{Llama2-70b}         &    23.1 & 11.5 & 38.9 & 0.0 & 23.2 & 11.6 & 38.8 & 0.0 & 25.0 & 13.3 & 39.9 & 0.0 & 25.2 & 12.6 & 41.7 & 0.0 & 25.4 & 12.7 & 42.0 & 0.0 \\ 
        \texttt{Llama3-70b}        &    \textbf{34.8} & \textbf{26.5} & \textbf{49.8} & 0.0 & 0.0 & 9.9 & 38.8 & 0.0 & \textbf{33.3} & 22.6 & 46.6 & 0.0 & 34.0 & 23.0 & 48.3 & 0.0 & 35.1 & 24.0 & 48.5 & 0.0 \\



      \bottomrule

    \end{tabular}

    }

    \caption{Summary of our results across nine autoregressive LLMs, open- and closed-source, organized by different classes of model size. Bin-f1 refers to the f1 score in the binary classification setting, where we only consider label 2 (\texttt{MisLC}) as the positive class. Ma-f1 and Mi-f1 are the macro- and micro-f1 for the 3-way classification task, where label 1 and 2 (\texttt{MisLC}, \texttt{Unclear}) are both positive classes. $\uparrow$ indicates higher is better, $\downarrow$ indicates lower is better.}

    \label{tab:baseline}


\end{table*}
\paragraph{Legal Database.} To align language models to our legal issues, we build a database using the full text of the documents compiled in Section \ref{sec:definition}.
We collect 27 documents with an average length of $\approx$ 24,000 tokens and the maximum being $\approx$ 96,000 tokens. Having such long documents in the database might cause a few problems: (i) the text chunks are significantly longer than the context window of some LLMs,
and 
(ii) most parts of the text chunk are irrelevant to the query. To this end, we perform a process to split the database into small, yet coherent, text chunks. Please refer to Appendix \ref{app:legaldb} for further processing steps.
\vspace{-1mm}
\paragraph{Web Search.} A crucial component of our legal tests is the availability of evidence $E_i$ for a piece of text $\mathbf{t}_i$. We query the Google Custom Search API\footnote{\url{https://developers.google.com/custom-search/v1/overview}} set to retrieve from the entire internet, using the same query we use for our legal database retrieval.
One issue is that web search does not return results if there are no sufficiently relevant findings --- we test 100 samples of our dataset and find this occurs for 26.5\% of FLARE queries and 37.9\% of RALM queries.
The web search returns various metadata such as the website link, the title, and the most relevant snippet from the webpage. We concatenate the snippets of the first result for each query and insert them into the prompt. We acknowledge this is not the most effective method --- there are many works on algorithms to iteratively retrieve evidence \cite{das_state_2023}. We urge further exploration of evidence gathering pipelines for future work.

\section{Experiment Results}
\label{sec:results}
\begin{table}[t]

\renewcommand\arraystretch{0.8}

    \centering

    \setlength{\tabcolsep}{2pt}

    \resizebox{0.85\linewidth}{!}{

    \begin{tabular}{p{4.5cm}|ccc}

    \toprule



          Setting&Bin-f1$\uparrow$ &Ma-f1$\uparrow$  & Mi-f1 $\uparrow$ \\
        \midrule
        \texttt{Random class} & 18.4 \small{$\pm 2.8$} & 17.4 \small{$\pm 1.8$} & 35.2 \small{$\pm 1.7$}\\
        \texttt{All label 2} & 23.1 & 11.6 & 38.8 \\
        \texttt{All label 1} & 0.0 & 9.9 & 38.8 \\
        \texttt{Mean Expert Performance} & 71.1 \small{$\pm 16.8$} & 64.9 \small{$\pm 16.7$} & 73.1 \small{$\pm 13.0$}\\
         \bottomrule

    \end{tabular}

    }

        \caption{Point-of-reference values for our binary and 3-way classification settings. Random class is a classifier where we sample predictions from a random distribution. The random class performance is taken over 100 runs. $\pm$ indicates standard deviation.}  

    \label{tab:ref_perf}

\end{table}

We perform experiments on a wide range of publicly available LLMs. Considering its importance for the legal domain and our task here, we extend our investigation to include Retrieval Augmented Generation (RAG).

\subsection{The State of the Art of LLMs on MisLC}
Our results are summarized in Table \ref{tab:baseline}. We also provide reference performances in Table \ref{tab:ref_perf}, where ``All label 2'' refers to the performance where every prediction is \texttt{MisLC}, ``All label 1'' refers to the performance where every prediction is \texttt{Unclear}, and ``Mean Expert Performance'' refers to the average human expert performance obtained from Section \ref{sec:dataset}. Bin-f1 refers to the f1 score in the binary classification setting, where we only consider label 2 as the positive class. Ma-f1 and Mi-f1 represent the macro- and micro-f1 for the 3-way classification task, where label 1 and 2 are separate positive classes, as defined earlier in this paper. Overall, the experiments show that the \texttt{MisLC} task is challenging for current large language models, even when augmented with retrieval, and they do not achieve human performance. This finding emphasizes the need to develop sophisticated methods to solve our \texttt{MisLC} task.
\paragraph{\texttt{MisLC} performance scales with general domain performance.} In general, the performance trends observed in \texttt{MisLC} align with models' general performance. For example, the open-source models \texttt{Mistral-7b} and \texttt{Solar-10b} are known to perform better than the default Llama-2 models\footnote{Based on general performance from the Hugging Face Open LLM leaderboard \url{https://huggingface.co/spaces/open-llm-leaderboard/open_llm_leaderboard}}, but the more recent Llama-3 models generally exhibit higher performance than others at similar sizes. The best performing closed-source model in the binary setting is \texttt{GPT-3.5-turbo}, performing +12 points f1 better than random guessing, while the best performing open-source model is \texttt{Llama3-70b} (+14.4 f1 score). For the 3-way classification, all the models except \texttt{GPT-4o} and \texttt{Llama3-70b} performed close to the random classifier baseline. This evidence suggests it is more challenging to predict the \texttt{Unclear} class than \texttt{MisLC}.
\paragraph{Larger models follow legal instructions more effectively.}
Older language models, especially the Llama 2 series, show high error rates (ER), i.e., failing to provide an expected keyword for 20-60\% of the answers. Upon inspecting the generations, we find they often \textit{refuse to answer the prompt} despite our prompt instructing otherwise. The balance between LLMs generating refusals and following instructions is constantly shifting in the field of AI Alignment (particularly red-teaming), so this might be an intentional shift in LLMs, but it might cause some concern in high-stakes domains. We also perform experiments without this constraint, allowing the model to generate freely and performing more extensive post-processing for evaluation. While the error rate decreases, the trends in performance are inconsistent. Please refer to Appendix \ref{sec:old_prompt} for further discussion on these additional experiments.

\subsection{Effect of Retrieval}
Our task is heavily reliant on external data, evidence $E_i$ and legal issues $L_i$, so a language model should be able to effectively retrieve and parse relevant knowledge. We retrieve from two sources: the legal resources used to create our definition, as described in Section \ref{sec:definition}, as well as web search. Similar to the above \textit{no-retrieval} setting, the models that have the best general domain performance benefit the most from retrieval. In particular, \texttt{GPT-4o} is the only model with a significant increase in performance (+$9.0$ Bin-f1) compared to other models. In the smaller models, combining the two sources \textit{hinders} performance. Compared to the \textit{no-retrieval} setting, \texttt{Mistral-7b} has a decrease in performance (-$11.2$ Bin-f1). Its 3-way classification performance remains constant, due to the model's improved performance on the \texttt{Unclear} class.

\begin{figure}[t!]%
  \centering
  \begin{subfigure}[c]{0.9\linewidth}
    \centering
    \includegraphics[trim={0.5cm 3.5cm 0.5cm 10.5cm},clip, width=\linewidth]{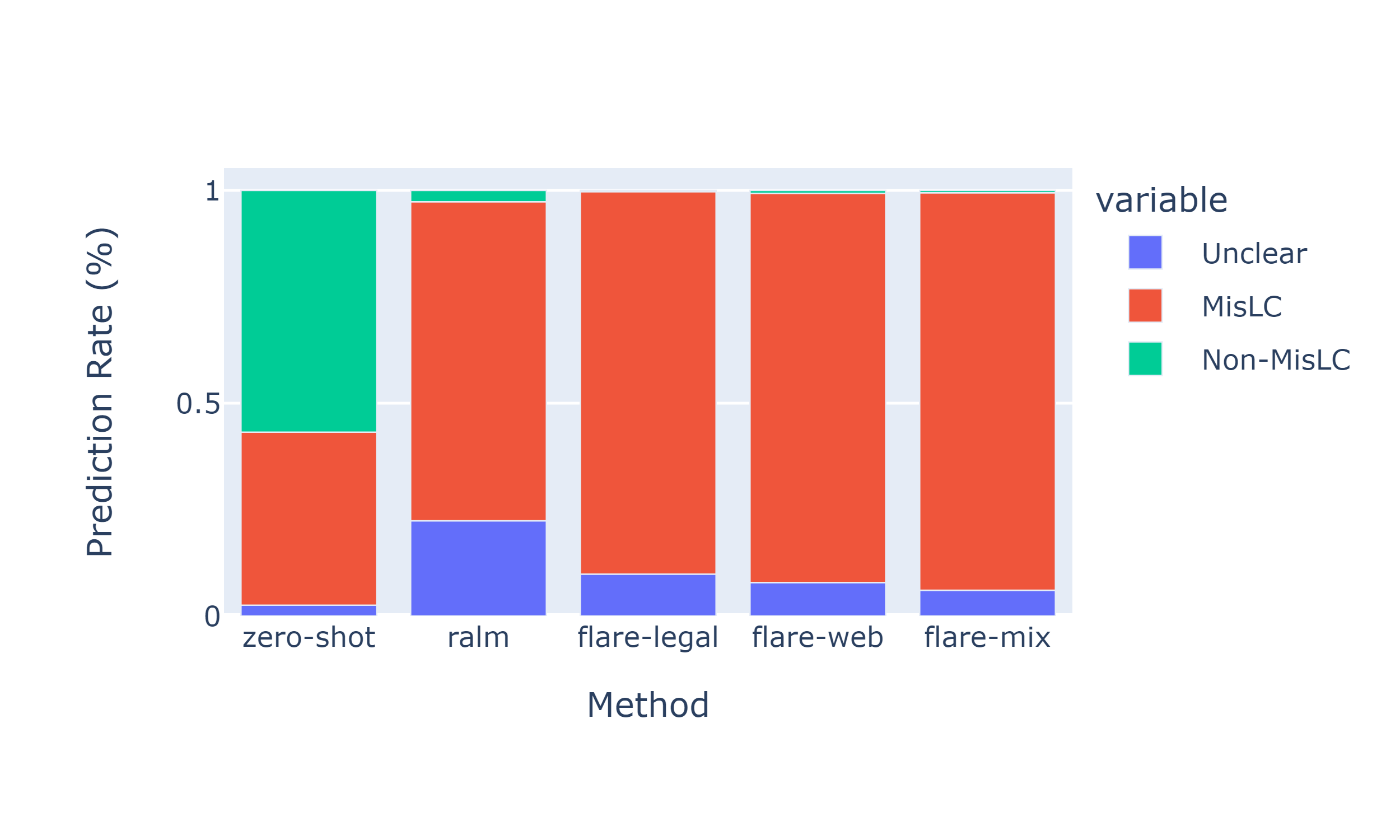}

    \caption{\texttt{Llama2-13b}.}
  \end{subfigure}
  \begin{subfigure}[c]{0.9\linewidth}
    \centering
    \includegraphics[trim={0.5cm 3.5cm 0.5cm 5.5cm},clip, width=\linewidth]{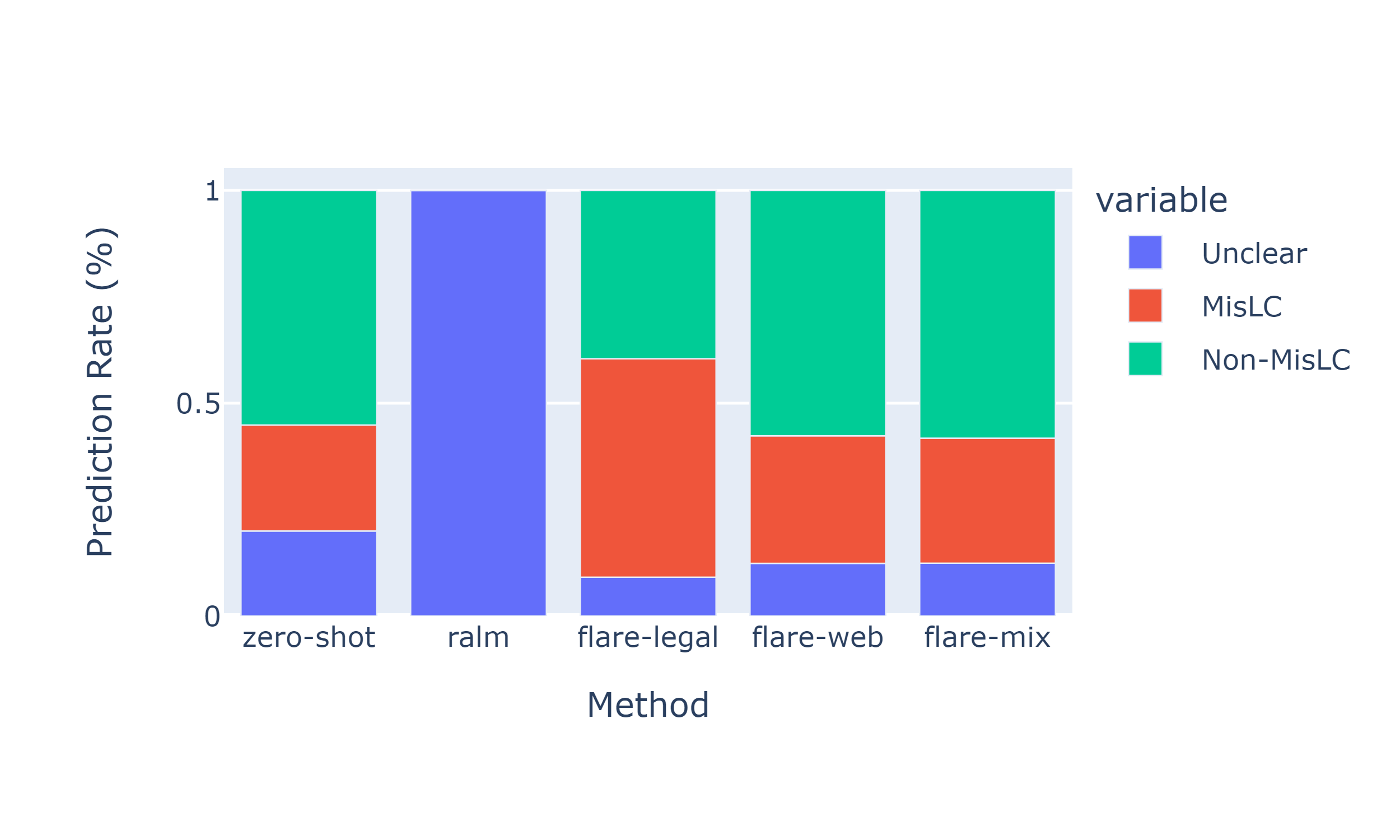}
    \caption{\texttt{Llama3-70b}.}
  \end{subfigure}
\setlength{\belowcaptionskip}{-8pt}
  \caption{Label distribution of the model predictions in our five settings for the best- and worst-performing models with retrieval.}
  \label{fig:plots}
\end{figure}

The label distributions of the best and worst models are shown in Figure \ref{fig:plots}, and other models can be found in Figure \ref{fig:dist_ext}. The worst-performing model \texttt{Llama2-13b} predicts the majority of our samples as \texttt{MisLC}, i.e. they tended to over-predict the positive class. Of the models tested, \texttt{GPT-4o} had the most consistent label balance across all experimental settings. It is important to note the positive class is relatively rare in our dataset; we provide reference values in Table \ref{tab:ref_perf}, but the reported performances is also heavily dependent on the distribution of predicted labels for each model.

Some models are not responsive to the retrieval methods combined with our task. For example, the \texttt{Llama 3} series predict only the \texttt{Unclear} class with the IC-RALM retrieval method, scoring $0.0$ points on Bin-f1. We hypothesize this is due to the frequency of retrieval in \texttt{IC-RALM} compared to \texttt{FLARE}; \texttt{FLARE} chooses when to retrieve adaptively based on the minimum model perplexity in a generated sentence. This indicates that retrieving too often can harm performance --- even in general domain tasks, FLARE's dynamic retrieval is found to perform better than static methods \cite{jiang-etal-2023-active}. We perform additional experiments to explore this hypothesis in Appendix \ref{sec:retrieval_ext}, and we urge further study in this direction.

\subsection{Detailed Analysis and Ablations}

While retrieval is important due to the broad range of knowledge required to detect and classify misinformation, we also examine the effectiveness of the models when directly given the legal issues $L_i$ and evidence $E_i$. We present two ablations with the FLARE pipeline: \textbf{Random-legal}, where we retrieve a random document from the legal dataset as a lower bound, and an \textbf{Oracle} setting as an upper bound. In the oracle setting, we provide the \textit{definition} of the ground truth legal issues $L_i$ as shown in Table \ref{tab:areas}. If there are no legal issues, we perform retrieval as per our normal pipeline. We also consider the ground truth evidence $E_i$, where we download the sources provided by legal annotators as HTML files, extract the first 500 characters of text, and concatenate all sources as the retrieved document.
We present results with \texttt{GPT-4o}, our best-performing model, as well as \texttt{Llama3-70b} (in Appendix \ref{app:ablations}).

As shown in Table \ref{tab:ablation-main}, the random document does not confuse the model, with performance increasing consistently by approximately 2 points f1 across all metrics. The oracle setting demonstrate improvement when only performing web search. We observe a decrease in performance when utilizing the ground truth definitions of our legal issues. This indicates the context afforded by the legal resources benefits model performance more than just a definition, but the retrieval algorithm does not necessarily choose the most relevant documents.



    












\begin{table}[t]

\renewcommand\arraystretch{1.1}

    
    \centering
    \setlength{\tabcolsep}{2pt}

    \resizebox{1\linewidth}{!}{

    \begin{tabular}{p{2.7cm}|ccc||p{3.0cm}|ccc}

    \toprule

GPT-4o &Bin-f1$\uparrow$ & Ma-f1$\uparrow$  & Mi-f1$\uparrow$ & Ablation &Bin-f1$\uparrow$ & Ma-f1$\uparrow$  & Mi-f1$\uparrow$  \\

        \midrule


    \texttt{FLARE}\small{(legal)}     &      32.3 & 25.8 & 46.7 &  \texttt{Random}\small{(legal)}    &   34.5 & 27.3 & 48.2 \\ 
      \texttt{FLARE}\small{(legal)}   &    32.3 & 25.8 & 46.7 &    \texttt{Oracle}\small{(legal)}     &  32.3 & 25.6 & 46.5 \\ 
     \texttt{FLARE}\small{(web)}       &    34.5 & 26.2 & 47.4 &   \texttt{Oracle}\small{(web)}      &   36.4 & 28.5 & 46.5 \\ 
      \texttt{FLARE}\small{(legal+web)} &   37.7 & 28.0 & 46.7 &   \texttt{Oracle}\small{(legal+web)} &  35.9 & 27.6 & 46.2 \\ 
      \bottomrule

    \end{tabular}

    }

   \caption{Summary of our ablations with \texttt{GPT-4o} using FLARE pipeline.}

    \label{tab:ablation-main}

\end{table}

\section{Conclusion}
We introduce a new task: Misinformation with Legal Consequence (\texttt{MisLC}) built on a body of literature spanning 4 broader legal topics and 11 fine-grained legal issues. A comprehensive study is performed on a wide range of open-source and proprietary LLMs that covers a broad parameter spectrum and varying training data. We also adapt existing works in Retrieval-Augmented Generation (RAG), retrieving from the web as well as our curated body of legal documents. 
We show the task remains challenging for the existing state-of-the-art large language models, even with the use of RAG. We also demonstrate trends from general domain tasks, such as a higher frequency of retrieval negatively impacting performance, is also reflected in performance on our specialized legal dataset. We hope our work can enable future research on this important task
with significant societal impact.

\newpage

\section*{Limitations}

\paragraph{Legal definitions.} As alluded to in various sections of the paper, misinformation is not its own legal issue. There are many historical cases where legal solutions to misinformation have been misused for censorship, and then repealed.\footnote{R. v. Zundel, [1992] 2 S.C.R. 731} Some argue the government should not be the arbitrator of the truth \cite{o2021perils}\footnote{\url{https://www.canada.ca/en/canadian-heritage/campaigns/harmful-online-content/summary-session-eight.html}}. However, the growing menace of online misinformation and disinformation underscores the urgent need for policy intervention. Regulation is an increasingly viable strategy, exemplified by the European Commission's recent action plan aimed at combating online disinformation.~\footnote{\url{https://digital-strategy.ec.europa.eu/en/policies/online-disinformation}}

\paragraph{Dataset size and composition.} We acknowledge the topic of the Russia-Ukraine conflict, and the range of legal issues found through our annotations is relatively limited. However, we are presenting a dataset that is reflective of a real-world use case, so attempts to artificially inflate the rarer legal issues would conflict with our motivations. We present all legal issues to demonstrate the comprehensiveness of our legal research, but the outcome of our annotations is meant to present the real-world distribution for our specific legal topic. 

We acknowledge our annotations and the retrieved evidence for determining the truthfulness of a statement were based on information available significantly after the development of the Russia-Ukraine conflict. However, we retain the ground truth evidence used by our legal experts, which we use in our Oracle setting in Section 5.3. We believe this point does not detract from the quality of our dataset, or our key conclusions, as the basis of the veracity verification is still clear in our annotations.

Finally, the overall dataset is relatively small. However, these are high quality annotations by legal experts with a significant amount of metadata, and we wanted to pace the annotations fairly without compromising quality. The dataset is meant to be a validation dataset rather than one used for training, as mentioned below in Intended Use. 

\paragraph{Implementation details.} There are minor details in our work that rely on closed-source API solutions, such as OpenAI Chat Completions API and Google Search, that reduces the reproducibility. Additionally, the adversarial filtering method we used has significant variance in the chosen samples every run. We did implement the AFLITE method used in WinoGrande \cite{sakaguchi2021winogrande} and found the difference between the two methods to be negligible after inspecting the samples manually. We will also posit that many models such as OpenAI specify they are not meant for domain-specific applications --- our results are meant to benchmark current performance and demonstrate there is continued room for improvement. Additionally, there is no legal LLM currently released, despite previous works calling for its development \cite{dahan2023lawyers}.

\section*{Ethics Statement}
\paragraph{Intended Use.} 
This paper defines a new task for harnessing misinformation societal harms and encourages researchers to develop more advanced algorithms to mitigate this. The dataset is meant to be a validation dataset rather than one used for training. Some applications include:
\begin{itemize}
    \item \textbf{Content Moderation for Social Media Platforms:} Social media platforms can use such a system to moderate content and identify misinformation that could potentially lead to legal liabilities. This can help platforms comply with regulations related to illegal content, defamation, hate speech, or other forms of harmful content.
    \item \textbf{Compliance Monitoring for Regulatory Bodies:} Regulatory bodies responsible for overseeing social media activities can utilize such a system to audit compliance with laws and regulations related to online content. For instance, it can help identify posts that violate consumer protection laws, election regulations, or intellectual property rights.
    \item \textbf{Journalistic Integrity Verification:} News organizations can use the system to verify the accuracy of social media content before reporting on it. This can help uphold journalistic integrity and avoid publishing false information that could lead to defamation lawsuits or damage the credibility of the news outlet.
\end{itemize}
\section*{Acknowledgements}
The research is in part supported by the NSERC Discovery Grants and the Research Opportunity Seed Fund (ROSF) of Ingenuity Labs Research Institute at Queen's University.
\bibliography{anthology,custom}

\appendix

\section{Detailed Related Work}
\label{app:drw}
Misinformation is a serious issue with significant societal impact, as factual dissonance can cause disorder in peoples' worldviews \cite{doi:10.1080/02650487.2019.1586210}. One option to minimize the effect of misinformation is automatic regulation or content filtering. 
Automatic methods play an important role in detecting misinformation, as they can reduce manual labour costs in searching for emerging rumours \cite{das_state_2023}. In practice, many such automatic systems often result in a poor user experience due to their lack of transparency \cite{algorithmic-content-moderation}. 
There have been various works that address separate components of the fact-checking pipeline: identifying checkworthy claims, gathering sources on those claims, and cross-checking the sources to confirm veracity \cite{das_state_2023}. There is growing interest in how to address the problem with LLMs \cite{chen2023combating, bang2023multitask}, and emerging works proposing new methodologies for fact-checking \cite{pelrine2023towards, pan-etal-2023-fact}.  However, these works do not consider issues in the law. While there are concerns with regulating misinformation with the law \cite{o2021perils}, we argue it is because of this discourse that the laws that currently exist have undergone rigorous vetting processes and are balanced to reduce societal harm.
 The most similar work to ours in objective is \citep{luo-etal-2023-legally}, which finds there are discrepancies between hate speech detection works and the law. 

Generative models 
have recently demonstrated strong proficiency in a wide variety of tasks such as relevance, stance, topics, and frame detection in tweets \cite{gilardi2023chatgpt}. 
Many new methods have emerged following the success of RLHF, including Direct Preference Optimization (DPO) to train a policy directly into a language model \cite{rafailov2023direct}. There is also a wide breadth of literature on improving the reasoning of an LM. \cite{wei2022chain} introduced few-shot chain-of-thought (CoT) prompting, which prompts the model to generate intermediate reasoning steps before reaching the final answer.
Due to the success of CoT prompting and the quality of the reasoning, several newer models incorporate step-by-step demonstrations in the training process \cite{lightman2023let}. This can act as a form of knowledge distillation when a larger language model generates higher quality demonstrations for a smaller model \cite{mukherjee2023orca}.

Large Language Models (LLMs) have also demonstrated the ability to capture and memorize a vast amount of world knowledge during pretraining \cite{guu2020realm}. However, this knowledge is stored implicitly within their parameters, leading to a lack of transparent source attribution for the facts and information generated in their outputs \cite{rashkin2023measuring, manakul2023selfcheckgpt}. LLMs are also susceptible to hallucinations, potentially fabricating facts and sources in their responses \cite{ye2023cognitive}. 
While some previous works refer to these errors as hallucinations \cite{luo-etal-2023-legally}, more recent works clarify hallucinations as a plausible answer with fabricated facts \cite{ye2023cognitive}. One viable strategy to address factual accuracy is Retrieval-Augmented Generation (RAG), where the language model is given explicit knowledge from external corpora \cite{du-ji-2022-retrieval}. 
Broadly speaking, various RAG strategies differ in three aspects: i) retrieval as text chunks, tokens, or other text snippets, ii) how to integrate the retrieved text with the LM, and iii) when to trigger retrieval \cite{retrieval-lm-tutorial}. 
Some approaches prepend retrieved documents in the input layer of the LM, leaving the LM architecture unchanged \cite{guu2020realm,shi2023replug}. 
In this category, In-Context RALM (IC-RALM) \cite{ram-etal-2023-context} and FLARE \cite{jiang-etal-2023-active} methods do not require pretraining or fine-tuning LMs, which can be expensive due to the large LM sizes. 
RAG can also be done by incorporating the retrieved text in intermediate layers \cite{pmlr-v162-borgeaud22a,fevry-etal-2020-entities,dejong2022mention}, or the output layers \cite{Khandelwal2020Generalization, he-etal-2021-efficient}. These approaches require access to the intermediate layers of the models, changes to the LM architecture, and/or further training in order for the model to use the data effectively.

\section{Additional Dataset Details}
Please refer to Table \ref{tab:samples} for example social media posts from our dataset.

\begin{table}[t]
\renewcommand\arraystretch{0.75}
    \centering\resizebox{\linewidth}{!}{
    \begin{tabular}{p{2cm}|p{5cm}}
    \toprule
    Checkworthy \& \texttt{MisLC} & ``We can deploy troops halfway around the world, in the middle of the Iraq desert, and feed them lobster on Sunday night. The Russians can't even supply their troops 50 miles from their homeland with unexpired MREs.'' \\
    \midrule
    Checkworthy but not \texttt{MisLC}   & Some Russian performing artists are speaking out against Putin - NPR\\
    \midrule
    Not Checkworthy, not \texttt{MisLC} & ``Ukraine is my home. Every street, corner, alleyway, nook and cranny all over the country have made me what I am today. If all of that is lost  I have no idea who I'll be.''\\
    \bottomrule

    \end{tabular}}
    \caption{Cherry-picked samples from our dataset comparing crowd-sourced labels of Checkworthiness to our expert annotations of \texttt{MisLC}.}
    \label{tab:samples}
\end{table}

\subsection{Data Processing}
\label{app:data}
The dataset contains a year's worth of tweet metadata from February 2022 to February 2023, collected to facilitate further research in misinformation. We hydrated 1 million English-language tweets, from which 10,000 tweets are randomly sampled. This was performed in February 2023, before Twitter's API policy changes were enacted. We then used Google Translate’s language detection function\footnote{\url{https://cloud.google.com/translate/docs/reference/rest}} as a secondary filter for tweets exclusively in English. All usernames (words starting with an @ symbol) in the tweets are replaced with <user>, and we remove unicode characters by encoding to ASCII. Finally, we identify social media posts with claims using a fine-tuned version of DeBERTa for claim detection\footnote{\url{https://huggingface.co/Nithiwat/mdeberta-v3-base_claim-detection}}, stopping once we have 4,000 samples. We also tested ChatGPT, but find DeBERTa is better aligned to Claim vs. No Claim annotations by our research team, which was performed based on previous definitions from \cite{das_state_2023}.

\subsection{Retrieval Database Preprocessing}
\label{app:legaldb}
We convert the text from PDF to HTML format using Adobe Acrobat, and then split each document into paragraphs by searching for two consecutive newline characters. Next, we rejoin the paragraphs in chunks of 2048 tokens with a 50\% sliding window context to preserve one paragraph's context and relationships with its immediate neighbours. After the splitting process, we obtain 590 text chunks. We build a positional BM25 index upon them using Pyserini \cite{Lin_etal_SIGIR2021_Pyserini}.

\subsection{Crowd-sourced Annotations}
\label{app:annotation}

Please refer to Table \ref{tab:instructions} for crowd-sourced annotation instructions. We first chose workers on Mechanical Turk through a prescreening process. We sampled 100 tweets and collected a set of annotations from two researchers given the instructions in Table \ref{tab:instructions}. The researchers' labels had a fourth option of ``ambiguous'' --- that is, these samples appeared to be too subjective to indicate good understanding of the worker's performance. This ``ambiguous'' label is automatically assigned where the researchers disagreed, or if one researcher preemptively assigns a sample as ambiguous. Then, we scored all workers with the researcher annotations as a ground truth. A worker needed to have a 70\% agreement with researcher annotations, excluding ambiguous samples, and they needed to have completed at least 10 HITs in the prescreening to be considered for further annotation. Among the annotators that met all requirements, two of them only labelled ambiguous samples --- for them, we sent a secondary test to obtain a fair assessment. We compensated the workers at \$0.18 per HIT.

\subsection{Adversarial Filtering} 
\label{app:filtering}
We use embeddings from RoBERTa-Large, and train the linear classifier with a KL divergence loss objective as shown in Equation \ref{eq:loss}. Since this is a different task from binary classification, we do not set a fixed $\tau$ --- instead, we take $\tau$ to be the mean loss over the entire dataset, which we indicate with $\tau_\mu$. During our filtering process, we find $\tau_\mu$ to be approximately 0.1 across three rounds.

\section{Experiment Details}
\label{app:settings}

\subsection{Model Choice}
\label{app:models}
\begin{itemize}
    \item \texttt{GPT-3.5-turbo}  \cite{ouyang2022training} --- A closed-source model trained with Reinforcement Learning with Human Feedback (RLHF). We performed experiments in June of 2024.
    \item \texttt{GPT-4o}\footnote{\url{https://openai.com/index/hello-gpt-4o/}} --- A closed-source model trained with Reinforcement Learning with Human Feedback (RLHF). We performed experiments in June of 2024.
    \item \texttt{Llama2-(7b, 13b, and 70b)} \cite{touvron2023llama} --- A suite of open-source models trained using RLHF, as well as safety fine-tuning to enhance helpfulness.
    \item \texttt{Llama3-(8b, and 70b)}\footnote{\url{https://ai.meta.com/blog/meta-llama-3/}} --- A suite of open-source models trained using a combination of supervised fine-tuning (SFT), rejection sampling, proximal policy optimization (PPO), and direct preference optimization (DPO), with a focus on safety fine-tuning to enhance helpfulness.
    \item \texttt{Mistral-7b} \cite{jiang2023mistral} --- A model trained with instruction tuning; rather than reinforcement learning, they fine-tune directly on instruction data.
    \item \texttt{Solar-10b}  --- This is a merged model that combines the instruction-tuned version of Solar, Solar-Instruct \citep{kim2023solar}, and OrcaDPO, another checkpoint trained with Direct Preference Optimization \cite{rafailov2023direct}. These two models are merged with Spherical Linear Interpolation (SLERP) \cite{barrera2004incremental}.
\end{itemize}
We choose Llama 2 and 3 to isolate the effect of model size, since these suites of models are trained with the same method at various parameter counts. We compare this suite to three open-source models trained on various combinations of fine-tuning, instruction tuning, and Proximal Policy Optimization (PPO) or Direct Policy Optimization (DPO).
We compare them to Mistral and Solar as they are the best-performing models on the Huggingface Open LLM leaderboard.\footnote{\url{https://huggingface.co/spaces/HuggingFaceH4/open_llm_leaderboard}} We searched for the best model in three sizes (7B parameters, $\approx$13B parameters, and 60B+ parameters) as of January 2024.  The \texttt{Solar-10b} we test combines two checkpoints of Solar \citep{kim2023solar}: Solar-Instruct, trained with instruction tuning, and OrcaDPO.

Additionally, we compare these against the closed-source \texttt{GPT-3.5-turbo} and \texttt{GPT-4o} as an alternative for limited computational resources. 
We utilize OpenAI's Chat Completions API for ChatGPT, and Hugging Face's text generation pipeline for all other models. For all models, we choose a sampling temperature of 0.3 after a hyperparameter search, testing temperatures of 0.1, 0.3, 0.5, 0.8, and 1.

\subsection{Prompting}
\label{app:prompting}
Please refer to Table \ref{tab:template} for our prompting format. While individual model prompts might vary based on their specific template formatting requirements, the core text is held constant throughout all of our experiments.

\begin{table}[t]
\renewcommand\arraystretch{0.75}
    \centering\resizebox{\linewidth}{!}{
    \begin{tabular}{p{2cm}|p{5cm}}
    \toprule
    Retrieval & Here is some relevant legal context on ``misinformation'': [doc] \newline \newline Web search results for the claim: [snippets]\\
    \midrule
    Classification    &Claim: [claim] \newline Classify the claim as either ``factual'' or ``misinformation.'' 
    
\\
    Constraints & Do not refuse to answer. Do not engage in explanations and politeness. Only respond with the words ``misinformation'', ``factual'', or ``unsure''. Do not add further context. \\
    \midrule
    \bottomrule

    \end{tabular}}
    \caption{Prompt template used in our experiments. We use a simple sentence to indicate the context of our retrieved document [doc] and/or web search results [snippets], and a keyword `Claim' to indicate the target input text within the prompt. In the main results, we also add some instructions to constrain the output to a single keyword.}
    \label{tab:template}
\end{table}

\subsection{Retrieval}
\label{app:retrieval}
\begin{figure}[t!]%
  \centering
  \begin{subfigure}[c]{\linewidth}
    \centering
    \includegraphics[trim={0.20cm 2.5cm 0.25cm 0cm},clip,width=\linewidth]{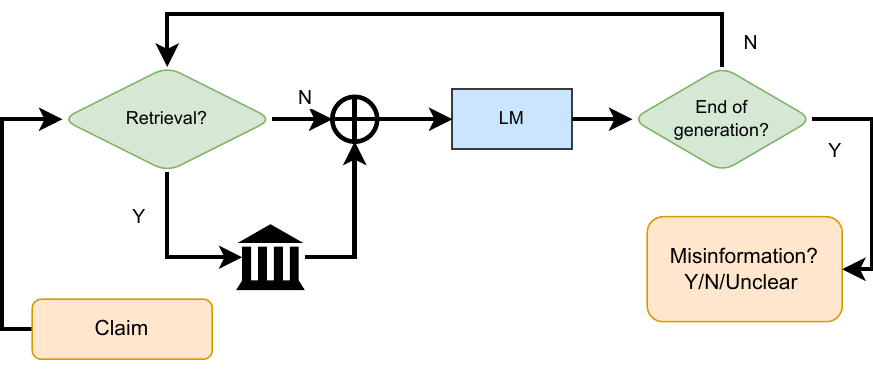}
    \caption{IC-RALM.}
    \label{fig:plotft}
  \end{subfigure}
  \begin{subfigure}[c]{\linewidth}
    \centering
    \includegraphics[width=\linewidth]{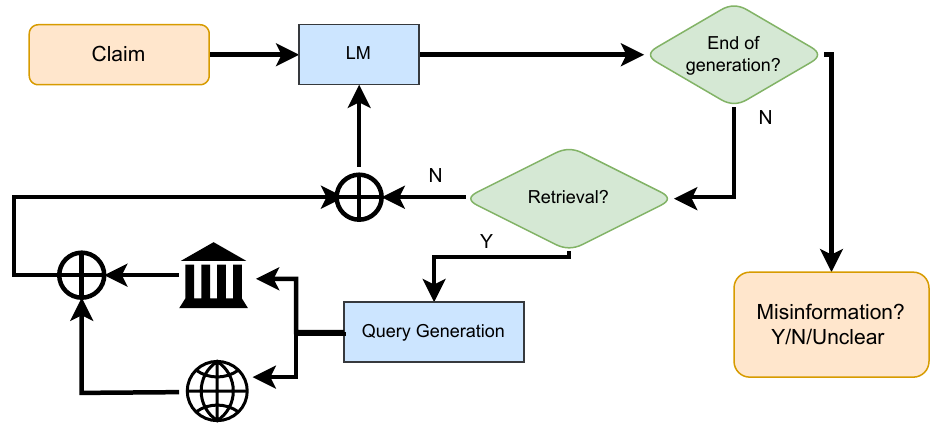}
    \caption{FLARE.}
  \end{subfigure}
  \caption{Illustrations of the IC-RALM and FLARE retrieval architectures.}
  \label{fig:retrieval}
\end{figure}

Please refer to Figure \ref{fig:retrieval} for an illustration of the two architectures. We take the implementation of IC-RALM directly from their Github,\footnote{\url{https://github.com/ai21labs/in-context-ralm}} and take most of FLARE's original implementation\footnote{\url{https://github.com/jzbjyb/flare}} except for the generation. We use ChatGPT for query generation in FLARE --- we also tested query generation using the same model (i.e. generating queries with \texttt{Llama3-70b}, retrieving, and then generating the final answer with \texttt{Llama3-70b}) and found performance comparable. 

\subsection{Hyperparameter tuning and Hardware Specifications}
\label{app:hyperparams}

\begin{figure}[t]
    \centering
    \includegraphics[width=\linewidth]{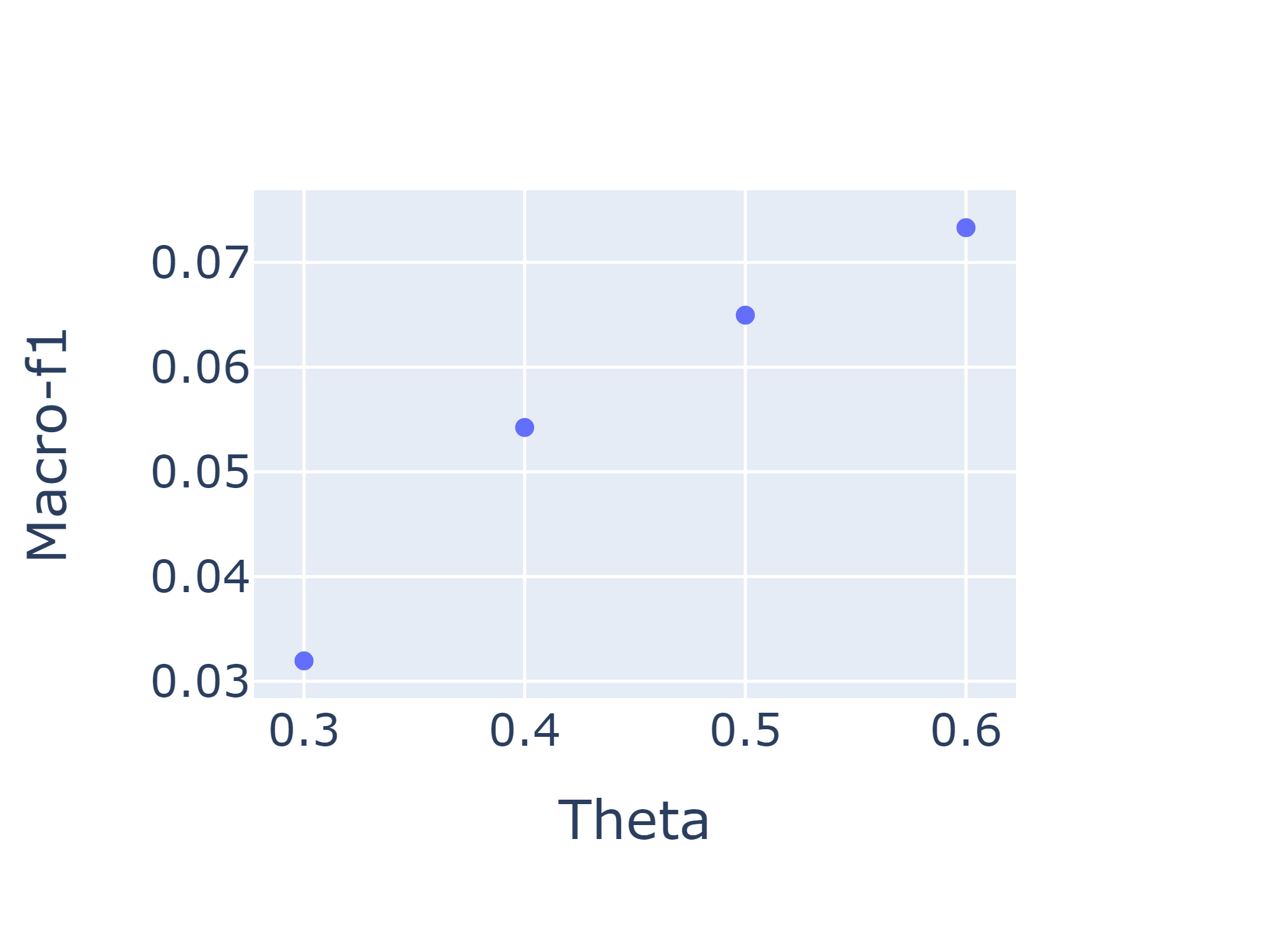}
    \caption{Change in macro-f1 as we increase $\theta$ over the first 100 samples.}
    \label{fig:hyperparameter}
\end{figure}

For the IC-RALM experiments, we set the \textit{stride} parameter to the $s = 4$ tokens that was used in most of \cite{ram-etal-2023-context} experiments, as it keeps a balance between performance and efficiency. This parameter is the frequency of which the retrieval is triggered. In FLARE experiments, we set the $\beta$ (the confidence threshold for query generation) value to be $0.4$ and did a grid search for $\theta$ (the confidence threshold for triggering retrieval) with $100$ samples of our dataset to find the best-performing value. We found performance scales consistently with $\theta$, as shown in Figure \ref{fig:hyperparameter}, and we choose $\theta = 0.5$ to balance performance with throughput.

We generate outputs with the vLLM library\footnote{\url{https://docs.vllm.ai/en/latest/}}, setting a maximum generation length of 1024. Experiment run times depended largely on the model size and experimental setting; smaller models took approximately 1.5 hours on our full dataset in the Flare (legal+web) setting, while larger models could take 3 hours. This equates to 1.5 GPU hours for smaller models, or 12 GPU hours for larger models. We conducted experiments with open-source models on a server cluster with a combination of Nvidia RTX6000-48GB and A100-40GB GPUs.
\section{Additional Experiments}
\begin{figure*}[t!]
  \centering
  \begin{subfigure}[c]{0.4\linewidth}
    \centering
    \includegraphics[trim={0.5cm 3.5cm 0.5cm 10.5cm},clip, width=\linewidth]{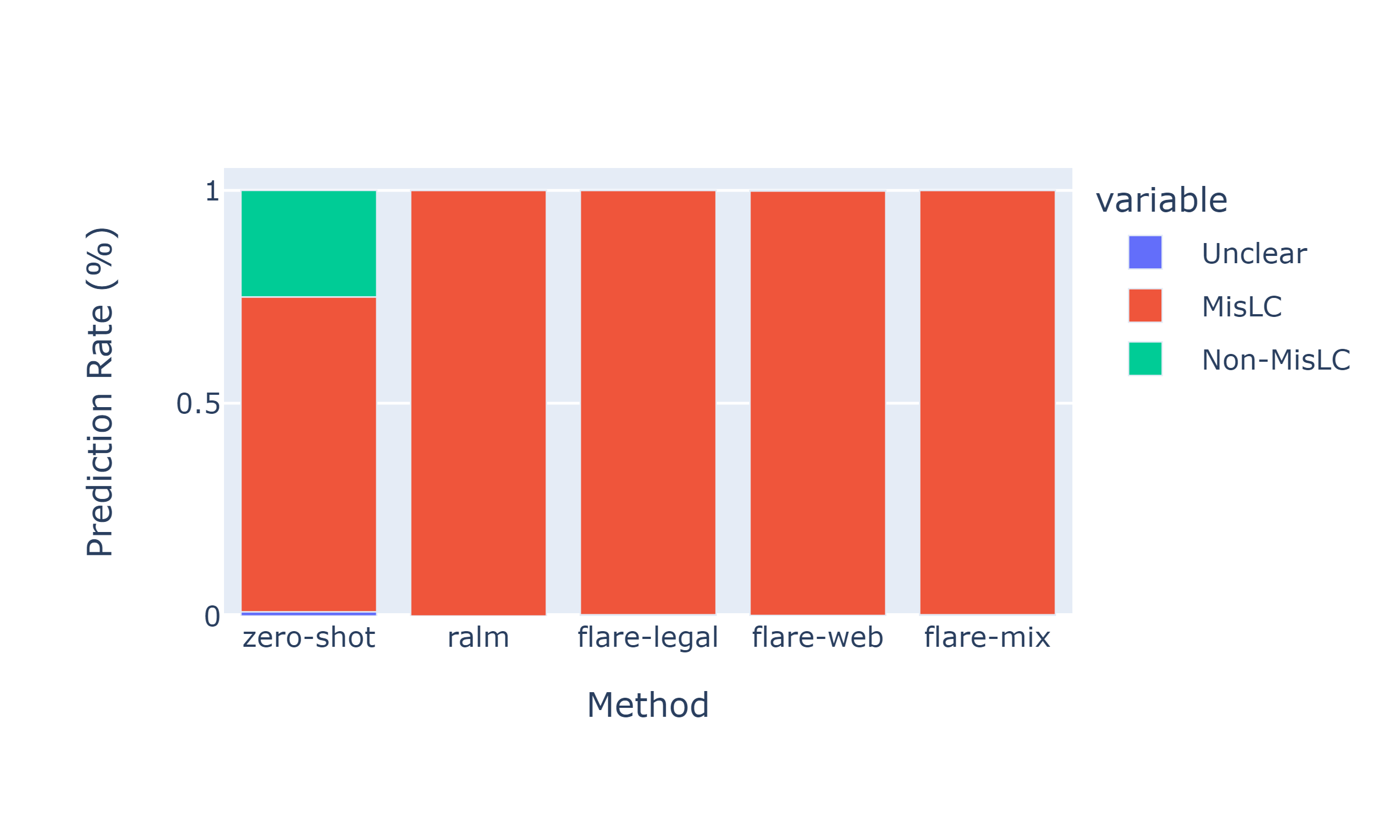}

    \caption{\texttt{Llama2-7b}.}
  \end{subfigure}
  \begin{subfigure}[c]{0.4\linewidth}
    \centering
    \includegraphics[trim={0.5cm 3.5cm 0.5cm 5.5cm},clip, width=\linewidth]{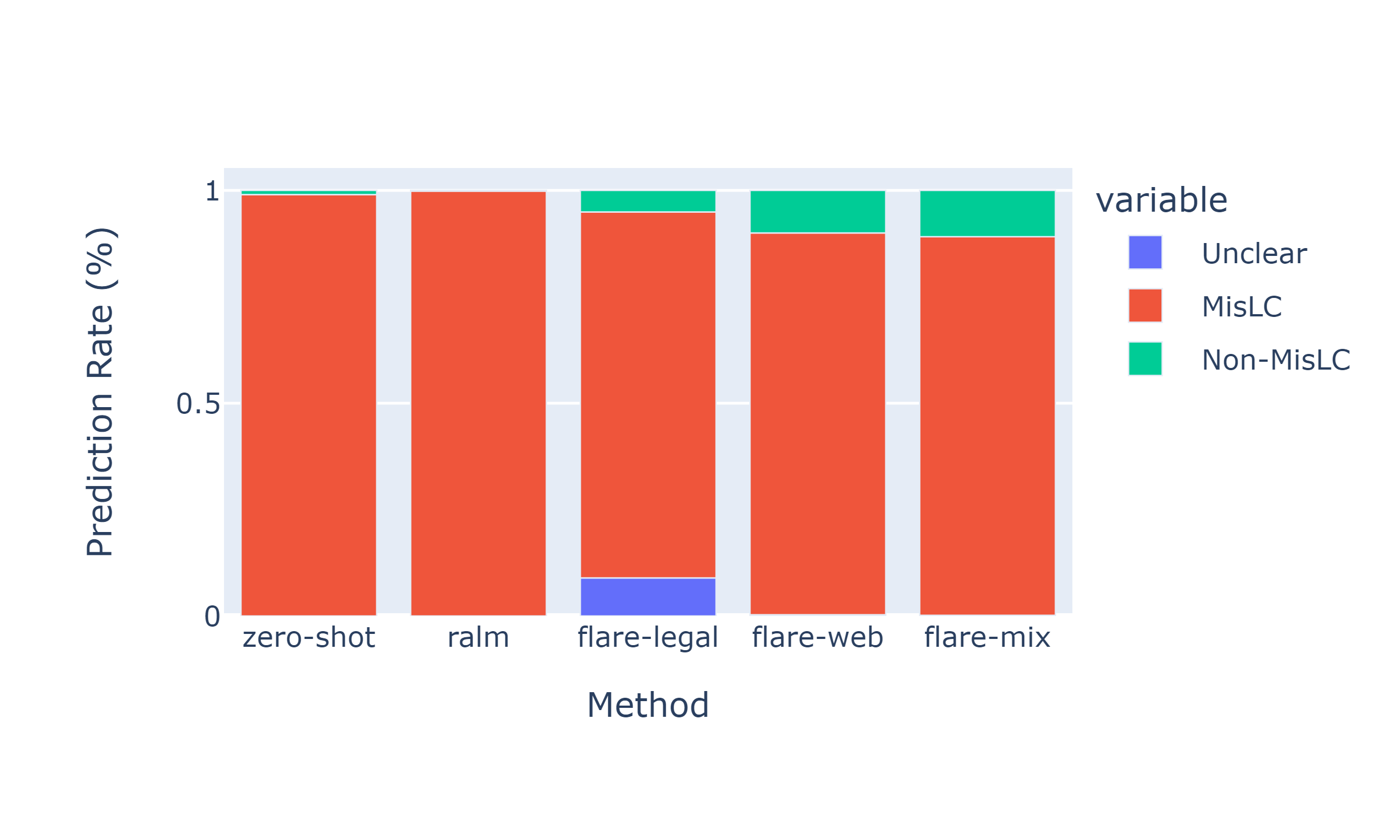}
    \caption{\texttt{Llama2-70b}.}
  \end{subfigure}
    \begin{subfigure}[c]{0.4\linewidth}
    \centering
    \includegraphics[trim={0.5cm 3.5cm 0.5cm 5.5cm},clip, width=\linewidth]{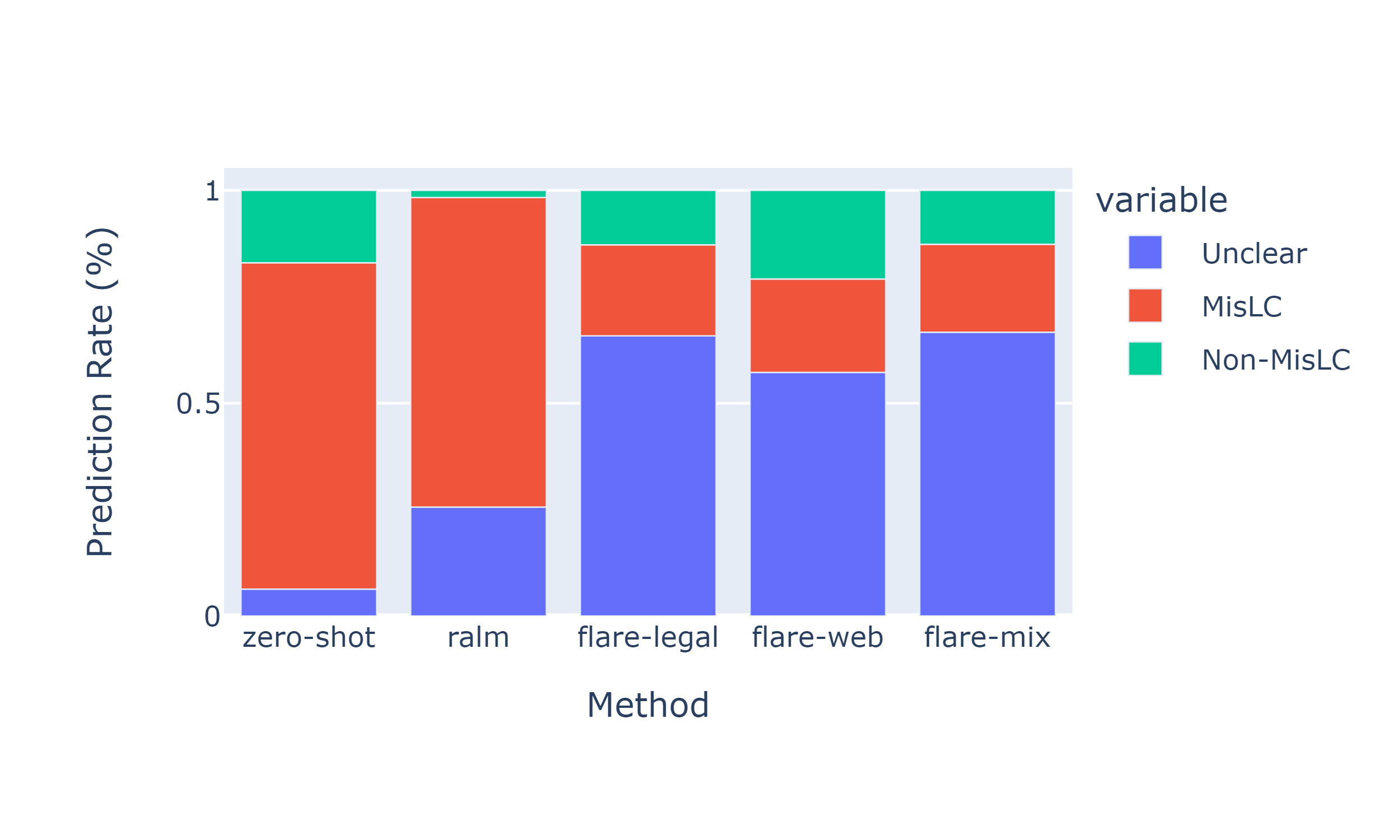}
    \caption{\texttt{Mistral-7b}.}
  \end{subfigure}
    \begin{subfigure}[c]{0.4\linewidth}
    \centering
    \includegraphics[trim={0.5cm 3.5cm 0.5cm 5.5cm},clip, width=\linewidth]{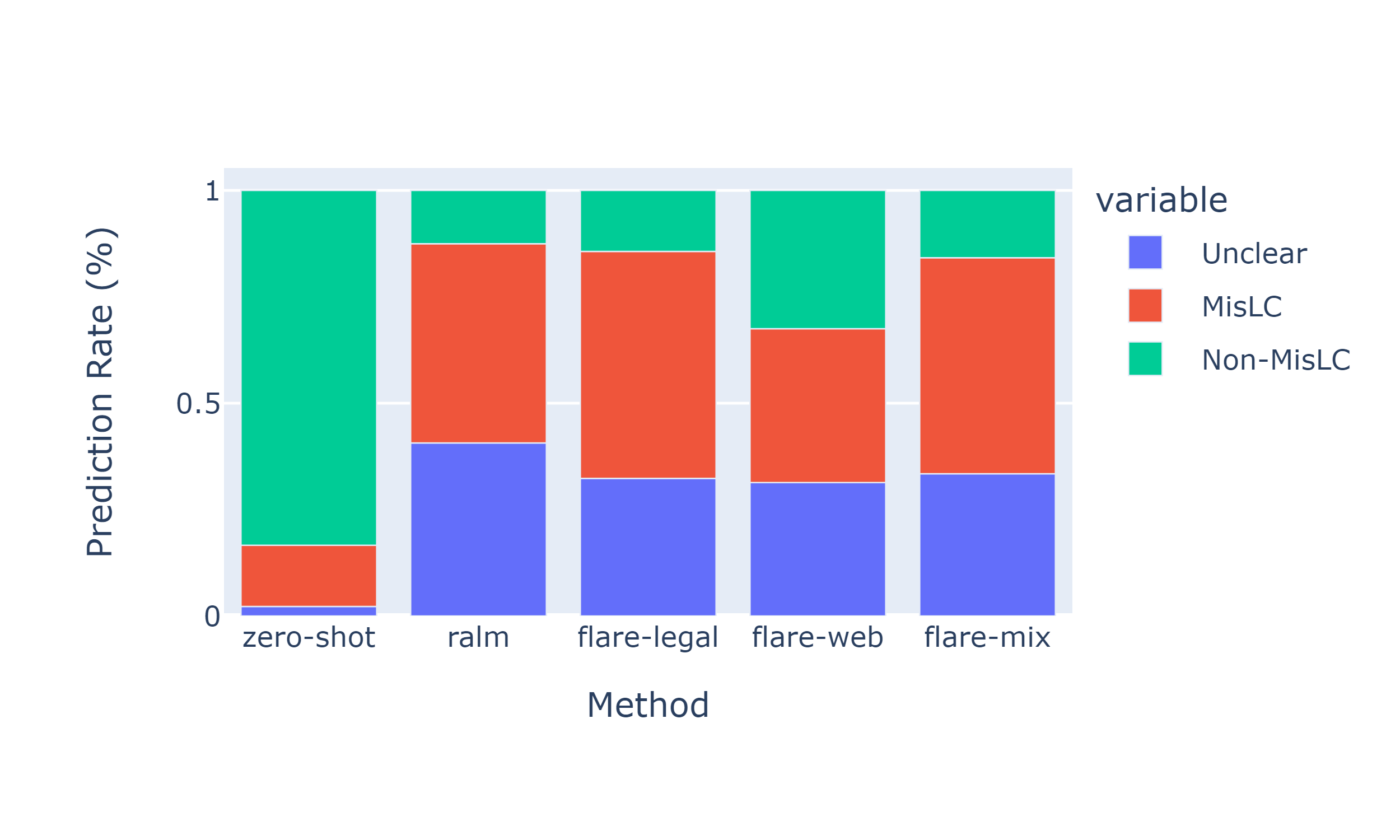}
    \caption{\texttt{Solar-10b}.}
  \end{subfigure}
    \begin{subfigure}[c]{0.4\linewidth}
    \centering
    \includegraphics[trim={0.5cm 3.5cm 0.5cm 5.5cm},clip, width=\linewidth]{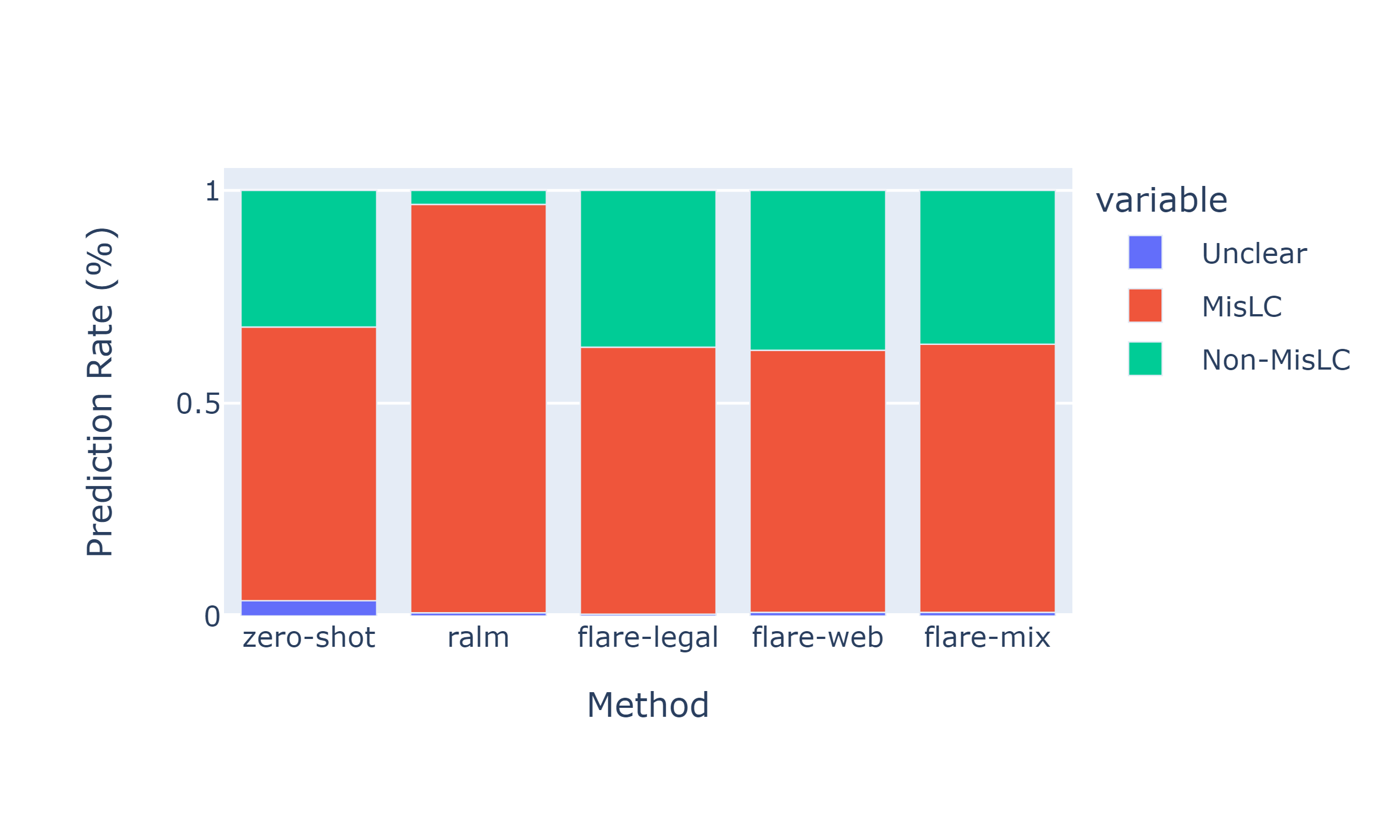}
    \caption{\texttt{GPT-3.5-turbo}.}
  \end{subfigure}
    \begin{subfigure}[c]{0.4\linewidth}
    \centering
    \includegraphics[trim={0.5cm 3.5cm 0.5cm 5.5cm},clip, width=\linewidth]{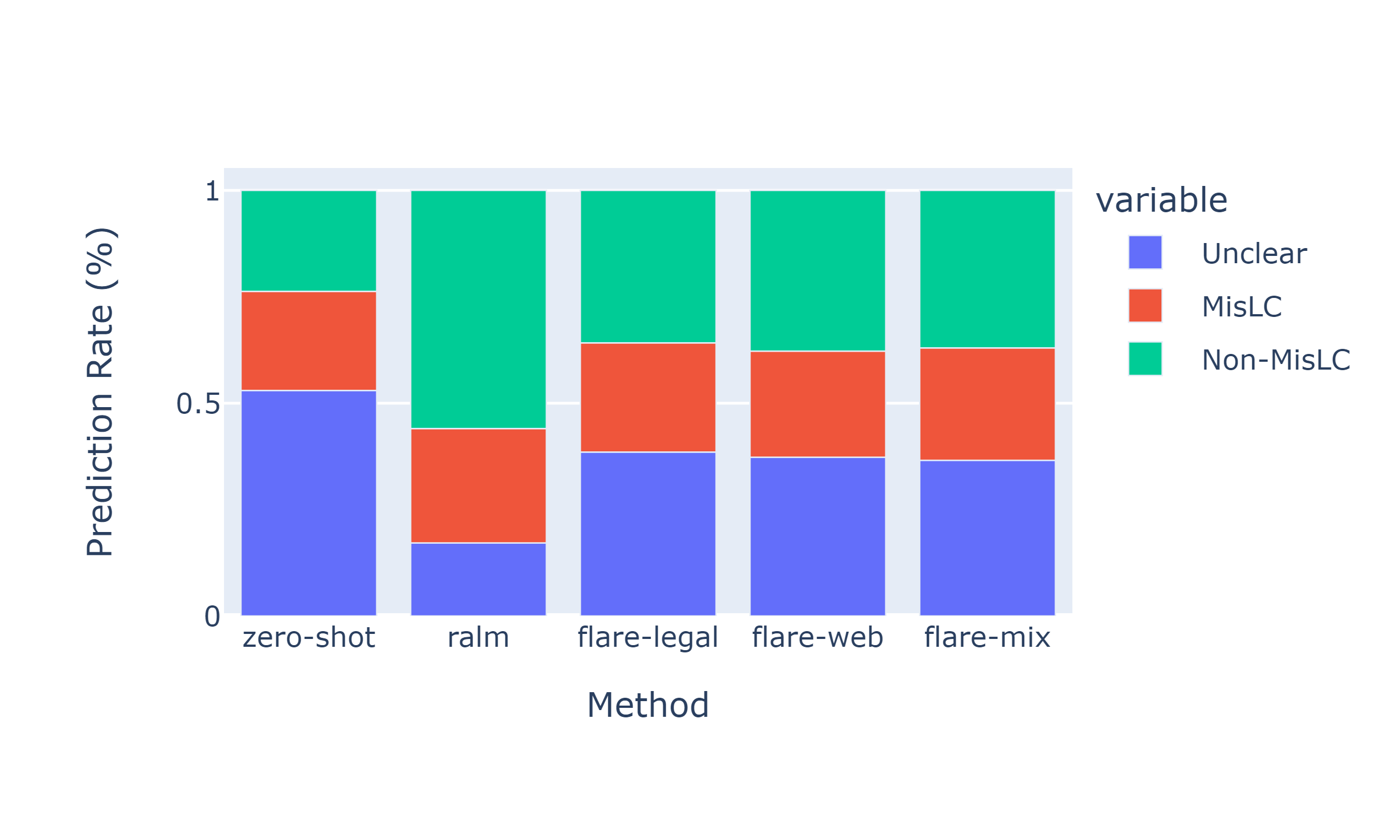}
    \caption{\texttt{GPT-4o}.}
  \end{subfigure}
      \begin{subfigure}[c]{0.4\linewidth}
    \centering
    \includegraphics[trim={0.5cm 3.5cm 0.5cm 5.5cm},clip, width=\linewidth]{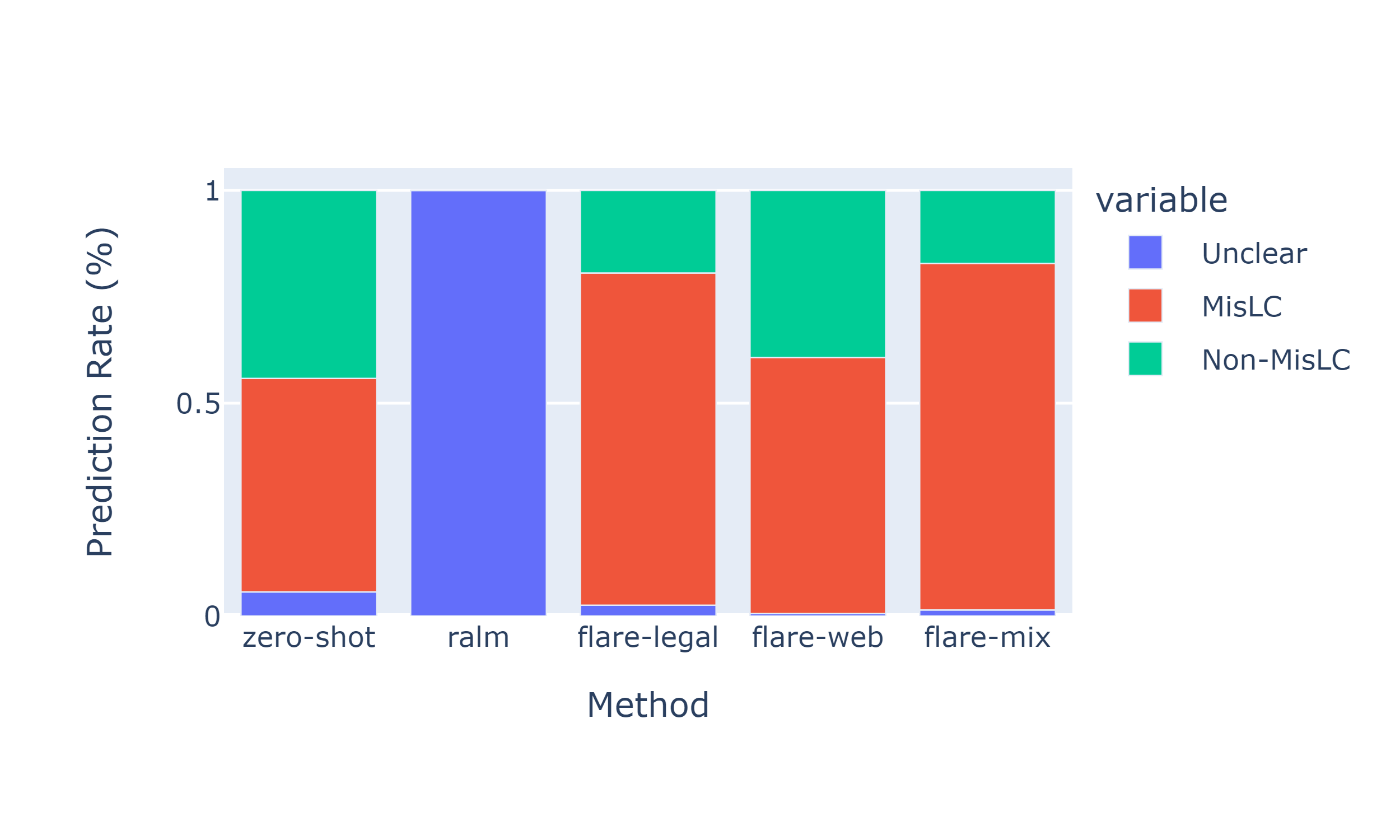}
    \caption{\texttt{Llama3-8b}.}
  \end{subfigure}
\setlength{\belowcaptionskip}{-8pt}
  \caption{Label distribution of the model predictions in our five settings for the remaining models tested.}
  \label{fig:dist_ext}
\end{figure*}

\begin{table}[t]

\renewcommand\arraystretch{0.8}

    
    \centering
    \setlength{\tabcolsep}{1.5pt}

    \resizebox{1\linewidth}{!}{

    \begin{tabular}{p{2.5cm}||cccc|cccc|cccc}

    \toprule

        \multirow{2}{*}{Model} &\multicolumn{4}{c|}{FLARE ($\theta=0.5$)} & \multicolumn{4}{c|}{FLARE ($\theta=1$)} & \multicolumn{4}{c}{IC-RALM (Legal)}\\

        \cmidrule{2-13}
        &Bin-f1$\uparrow$ &Ma-f1$\uparrow$  & Mi-f1 $\uparrow$ & ER $\downarrow$&Bin-f1$\uparrow$ & Ma-f1 $\uparrow$ & Mi-f1 $\uparrow$  & ER $\downarrow$&Bin-f1$\uparrow$  &Ma-f1 $\uparrow$ & Mi-f1 $\uparrow$ & ER $\downarrow$\\

        \midrule

        \texttt{GPT-4o} & 32.3 & 25.8 & 46.7 & 0.0 &30.7 & 25.0 & 46.7 & 0.0  &35.8 & 28.5 & 48.9 & 0.0\\
        \midrule
        \texttt{Llama3-8b} & 27.3 & 13.6 & 43.0 & 0.0  & 25.3 & 13.9 & 41.4 & 0.0  & 0.0 & 9.8 & 38.7 & 0.0 \\
        \texttt{Llama2-13b} & 22.2 & 17.8 & 38.8 & 0.0  & 22.6 & 16.0 & 39.3 & 0.0  & 22.6 & 18.2 & 39.7 & 0.1 \\

      \bottomrule

    \end{tabular}

    }

    \caption{A comparison of the RALM retrieval method with FLARE set to retrieve at every possible step (i.e. $\theta=1$). We conducted experiments for all models but only present results for these three to illustrate the effect of retrieval.}

    \label{tab:retriev_max}


\end{table}
\begin{table}[tb!]

\renewcommand\arraystretch{0.8}

    
    \centering
    \setlength{\tabcolsep}{4pt}

    \resizebox{1\linewidth}{!}{

    \begin{tabular}{p{2.7cm}|ccc||p{3.0cm}|ccc}

    \toprule

Llama3-70b &Bin-f1$\uparrow$ & Ma-f1$\uparrow$  & Mi-f1$\uparrow$ & Ablation &Bin-f1$\uparrow$ & Ma-f1$\uparrow$  & Mi-f1$\uparrow$  \\

        \midrule


    \texttt{FLARE} \small{(legal)}     & 33.3 & 22.6 & 46.6 &    \texttt{Random} \small{(legal)}    &          32.2 & 22.7 & 47.9 \\ 
      \texttt{FLARE} \small{(legal)}   &   33.3 & 22.6 & 46.6&  \texttt{Oracle} \small{(legal)}     &         32.0 & 21.1 & 47.5  \\ 
     \texttt{FLARE} \small{(web)}       & 34.0 & 23.0 & 48.3 &   \texttt{Oracle} \small{(web)}      &          32.2 & 22.7 & 48.7  \\ 
      \texttt{FLARE} \small{(legal+web)} &  35.1 & 24.0 & 48.5 & \texttt{Oracle} \small{(legal+web)} &         34.2 & 21.6 & 48.0  \\ 
      \bottomrule

    \end{tabular}

    }

    \caption{Summary of our ablations with \texttt{Llama3-70b} using FLARE pipeline.}

    \label{tab:ablation}

\end{table}
\subsection{Prompt Constraint and Error Rate}
\label{sec:old_prompt}
Previous works have observed legal tasks with long contexts often lead to a model being more ``decisive'' \cite{luo-etal-2023-legally}. In our experiments, we note that adding retrieved text to the input context significantly reduces the error rate. This suggests there is some trade-off between the instruction complexity and the safety fine-tuning performed for the Llama 2 models. Llama 2's safety fine-tuning has been noted to be unstable and easily reversed with a few steps of parameter-efficient fine-tuning \cite{lermen2023lora}, and we hypothesize this instability is also causing the fluctuations in error rate.

In addition to the stricter prompting instructions reported in the main body, we also evaluate the models without constraining the outputs --- i.e. simply asking for a classification, as shown in Table \ref{tab:template}. We evaluate the generations by searching the entire generated text for the keywords ``factual'' or ``misinformation.'' We first check for the keywords in quotes (`` ''), as that is the format given in the prompt, and then we check for all other mentions if quotes do not exist. If a model's generated text contains both of these keywords, we count this as an \texttt{unclear} prediction. For any generation without either keyword, we first filter over all model responses to analyse the responses. Many of these answers are non-answers, such as ``As an AI language model, I am unable to provide a response.'' is a non-answer, or an error in the generation. We report the Error Rate (ER) alongside macro- and micro-f1 score.

\begin{table*}[t]
\renewcommand\arraystretch{0.8}

    
    \centering
    \setlength{\tabcolsep}{1.5pt}

    \resizebox{1\linewidth}{!}{

    \begin{tabular}{p{3cm}||cccc|cccc|cccc|cccc|cccc}

    \toprule

        \multirow{2}{*}{Model}  & \multicolumn{4}{c|}{No Retrieval}&\multicolumn{4}{c|}{IC-RALM (Legal)}& \multicolumn{4}{c|}{FLARE (Legal)} &\multicolumn{4}{c|}{FLARE (Web)}& \multicolumn{4}{c}{FLARE (Legal+web)}\\

        \cmidrule{2-21}
        &Bin-f1$\uparrow$ &Ma-f1$\uparrow$  & Mi-f1 $\uparrow$ & ER $\downarrow$&Bin-f1$\uparrow$ & Ma-f1 $\uparrow$ & Mi-f1 $\uparrow$  & ER $\downarrow$&Bin-f1$\uparrow$  &Ma-f1 $\uparrow$ & Mi-f1 $\uparrow$ & ER $\downarrow$&Bin-f1$\uparrow$  &Ma-f1$\uparrow$  & Mi-f1 $\uparrow$ & ER $\downarrow$&Bin-f1$\uparrow$ &Ma-f1$\uparrow$  & Mi-f1 $\uparrow$ & ER $\downarrow$\\

        \midrule

        \texttt{GPT-3.5-turbo} & \textbf{30.9} & 22.0 & \textbf{44.2} & 0.3 & 0.0 & 15.0 & 26.7 & 0.0 & 30.4 & 19.9 & 45.0 & 0.1 & 31.8 & 25.2 & 44.3 & 0.1 & 30.4 & 16.4 & 48.2  & 0.0 \\
        \midrule
        \texttt{Mistral-7b} & 21.1 & \textbf{22.5} & 41.9 & 0.3 & 21.0 & \textbf{21.3} & \textbf{43.1} & 0.1 & 23.7 & 22.6 & 42.4 & 0.1 & 12.2 & 16.4 & 42.4 & 0.0 & 11.8 & 15.1 & 41.1 & 0.0  \\
        \texttt{Llama2-7b} & 21.1 & \textbf{22.5} & 45.5 & 0.1 & 23.0 & 20.9 & 40.5 & 0.1 & 16.5 & 18.5 & 40.4 & 0.7 & 23.3 & 22.6 & 41.4 & 0.4 & 18.9 & 20.1  & 40.1 & 0.9 \\  
        \midrule
        \texttt{Solar-10b} & 19.2 & 18.3 & 39.4 & 0.3 & \textbf{26.6} & \textbf{21.3} & 36.5 & 2.1 & 25.3 & 21.7 & 39.5 & 0.7 & 25.3 & 21.7 & 39.5 & 0.7 & 26.2 & 20.7  & 38.8 & 1.2  \\
        \texttt{Llama2-13b} & 18.0 & 17.7 & 40.0 & 0.0 & 13.2 & 15.0 & 41.3 & 0.3 & 17.7 & 19.5 & 41.4 & 1.5 & 17.5 & 19.6 & 41.2 & 0.9 & 18.7 & 20.0  & 41.5 & 1.5  \\
        \midrule
       \texttt{Llama2-70b} & 24.1 & 21.0 & 43.5 & 0.1 & 23.6 & 21.2 & 42.0 & 0.0 & 22.8 & 20.7 & 42.1 & 3.1 & 21.3 & 21.8 & 43.4 & 3.0 & 21.9 & 21.1  & 43.5 & 3.3  \\



      \bottomrule

    \end{tabular}

    }

    \caption{Summary of the unconstrained results across seven LLMs, open- and closed-source, organized by model size. Since the size of ChatGPT is unknown, we present it at the top.}

    \label{tab:baseline_unconstrained}


\end{table*}

\subsection{Retrieval Extended}
\label{sec:retrieval_ext}
We conducted an ablation with FLARE where we always performed retrieval on the legal dataset (i.e. set $\theta$ to 1) and observed similar performance as RALM across all models, summarized in Table \ref{tab:retriev_max}. While we conducted experiments with all of the models, we only present three key results. First, \texttt{Llama3-8b} had a Bin-f1 score of 0.0 with the IC-RALM retrieval method. However, FLARE even at the highest retrieval level does not exhibit this behaviour.

\subsection{Ablations Extended}
\label{app:ablations}

Please refer to Table \ref{tab:ablation} for results with \texttt{Llama3-70b}. As shown, the trend is similar to what was observed with \texttt{GPT-4o}.  

\begin{table*}
\centering
\begin{tblr}{
  width = \linewidth,
  colspec = {Q[500]},
  hlines,
  vlines,
}

We aim to identify checkworthy claims. A claim is defined as "stating or asserting that something is the case, typically without providing evidence or proof." 
\newline \newline
Examples of claims:
"The Eiffel Tower is the tallest tower in the world"
"Michael Jackson was seen at the department store last week"
"My wife can't eat chocolate because she's allergic"
\newline \newline
Not claims: opinions, emotions, exclamations. For example:

"I think Snow White was an idiot,"

"My wife is so nice and I love her,"

"Wow! Amazing!"

If there is no claim, please indicate "Empty/no claim. (3)"
\newline \newline
Please choose "Checkworthy" (1) if you consider at least one claim in the statement to be checkworthy. Checkworthy is defined as:
Having the potential to influence/mislead people, organizations and countries. If you read this statement, it would influence your opinion of the topic.
Discussing a topic, or quoting a person capable of signficant social impact.**
\newline \newline
It might be checkworthy if you can answer "yes" to any of these questions:

Does it provoke an emotional response?

Does it make a bold statement on a controversial issue?

Is it an extraordinary claim?

Does it contain clickbait?

Does it have topical information that is within context?

Does it use small pieces of valid information that are exaggerated or distorted?
\newline \newline
For example: "Biden's Climate Requirements: Cut 90\% of Red Meat From Diet; Americans Can Only Eat One Burger Per Month" is a checkworthy claim because it suggests the President of the United States wants to regulate peoples' diets. Some might feel angry because it is outside Biden's jurisdiction, so it is important to fact-check this statement.
\newline \newline
Choose Not Checkworthy (2) if the claim is not checkworthy. Not checkworthy claims are at least one of the following:

Innocuous (eg. Ryan Renolds has six fingers on his right hand)

Based on common knowledge (eg. water is wet, a cough makes your throat sore)

Made solely based on private information (eg. I had a sandwich for lunch yesterday) \\
\end{tblr}

\caption{Instructions provided to crowd-sourced (Mechanical Turk) workers for identifying checkworthiness.}
\label{tab:instructions}
\end{table*}


\section{Additional Legal Details}
\label{app:legal}
Please refer to Tables \ref{tab:areas} for a comprehensive list of legal issues considered in our annotations. 

\begin{table*}
    \centering
    \setlength{\tabcolsep}{2pt}

    \resizebox{0.8\linewidth}{!}{
        \begin{tblr}{
          width = \linewidth,
          colspec = {Q[150]Q[150]Q[398]Q[300]},
          hlines,
          vlines,
        }
        \textbf{Broad Legal Topic} & \textbf{Legal Issue}  & \textbf{Key legal tests} & \textbf{Defences} \\ 
        Defamation & Defamation  &  1. Defamatory in Nature (in the sense that the things in question would tend to lower the plaintiff’s reputation in the eyes of a reasonable person)\newline 2. Publication (communicated to a third party)
                    &  1. Qualified Privilege\newline 2. Responsible Communications\newline 3. Fair Comment (assuming (a) the comment is on a matter of public interest; (b) the comment is based on fact; (c) the comment, though it can include inferences of fact, is recognizable as comment; and (d) any person could honestly express that opinion on the proved facts)                                                     
        \\
        Freedom of Expression & Freedom of Expression   &  1. The activity must be an expressive, i.e. must ``convey meaning'' (``It might be difficult to characterize certain day-to-day tasks, like parking a car, as having expressive content.``)\footnote{Irwin Toy Ltd. v. Quebec (Attorney General), [1989] 1 SCR 927, 58 DLR (4th) 577.} \newline 2.Is the government's purpose, or otherwise effect, to restrict expression of this meaning?       
                                &  1. Can establish the ``truth,'' eg. clinical evidence\newline 2. Non-intent, i.e. published misinformation without intent\footnote{Canadian Constitution Foundation v. Canada (Attorney General), 2021 ONSC 1224}
        \\
        \SetCell[r=2]{h} Criminal Laws & Cyberbullying &    If false/inaccurate information is being spread to harass or harm others, the spreader
        could face cyberbullying or harassment charges
        &  N/A                                                                                                      
        \\
        & Public Mischief &    Every one commits public mischief who, with intent to mislead, causes a
        peace officer to enter on or continue an investigation by \newline (a) making a false statement that accuses some other person of having
        committed an offence; \newline
        (b) doing anything intended to cause some other person to be suspected of
        having committed an offence that the other person has not committed, or
        to divert suspicion from himself; \newline
        (c) reporting that an offence has been committed when it has not been
        committed; or \newline
        (d) reporting or in any other way making it known or causing it to be made
        known that he or some other person has died when he or that other
        person has not died.
        & N/A                                                                                                       
        \\
        \SetCell[r=2]{h} Consumer Protection Laws & Food and Drugs Act & Spreading false and private information about someone without their consent can lead to
        privacy violation claims because it infringes upon their right to control their personal
        information and keep it private. & N/A \\
        & Data Privacy & Under the Food and Drugs Act, Health Canada is tasked with (among other things)
        monitoring misleading health claims and regulatory enforcement to address health risks \newline Among other things, food in Canada shall not be sold or advertised in a manner
        that is false, misleading or deceptive & N/A
        
        \\                                                                                                    
        \end{tblr}
    }
\end{table*}

\begin{table*}

    \centering
    \setlength{\tabcolsep}{2pt}

    \resizebox{0.8\linewidth}{!}{
        \begin{tblr}{
          width = \linewidth,
          colspec = {Q[150]Q[150]Q[450]Q[300]},
          hlines,
          vlines,
        }
        \textbf{Broad Legal Topic} & \textbf{Legal Issue}  & \textbf{Key legal tests} & \textbf{Defences} \\ 
        Consumer Protection Laws & Federal Competition Act &        The Commissioner of the Competition and the Department of Public Prosecutions can
        initiate actions to address misleading claims using either of the criminal [section 52(1)] or
        civil tracks \newline
        - All representations that are false or misleading in a material respect, in any form,
        are subject to the Competition Act \newline
        - If a representation could influence a consumer to buy or use the product or
        service advertised, it is material \newline
        - NOTE: Propaganda and advertising are usually based on real accounts,
        with an incomplete focus on parts that are favourable to a campaign \cite{facts_fakenews}. \newline
        - To determine whether a representation is false or misleading, the courts consider
        the "general impression" it conveys, as well as its literal meaning   &  N/A                                                                                                       
        \\
        \SetCell[r=4]{h} Other (i.e., not fitting within one of the broad legal topics above) & Election Laws & The Canada Elections Act has prohibited false or misleading statements, since
        2018, about electoral candidates if they are expressed during the election period
        with the intention of affecting the results of the election. The Election Modernization Act sets out important transparency and
        disclosure requirements for political advertising \cite{dawood2020protecting} & N/A
        \\
        & Intentional Infliction of Mental Suffering & This common law tort involves intentionally inflicting emotional distress through acts or
        words which results in emotional harm as visible, provable illness. \newline
        - The plaintiff must prove 1) Act (Statement need not be false, but speech must be extreme), 2) Intent (i.e. calculated to produce harm), 3) Injury (i.e. the plaintiff must have suffered actual harm; some injury in the
        form of psychological harm) & N/A
        \\
        & Hate Speech &Fake news affects society as a whole, whereas hate speech harms individuals or members
        of a specific group \cite{katevas2022legal} & N/A
        \\
        & Intellectual Property & Trademarks Act provides that no person shall “make a false or misleading statement
        tending to discredit the business, goods or services of a competitor”, nor “make use, in
        association with goods or services, of any description that is false in a material respect
        and likely to mislead the public as to” the character, quality, quantity or composition, the
        geographical origin, or the mode of the manufacture production or performance of the
        goods or services.& N/A
        \\                                                                                                    
        \end{tblr} 
    }
  \caption{Areas of law that can be used to indict misinformation published online.}
  \label{tab:areas}
\end{table*}

\end{document}